\documentclass{article} 
\usepackage{iclr2027_conference,times}

\usepackage{amsmath,amsfonts,bm}

\def\eqref#1{equation~\ref{#1}}

\def\1{\bm{1}}

\DeclareMathAlphabet{\mathsfit}{\encodingdefault}{\sfdefault}{m}{sl}
\SetMathAlphabet{\mathsfit}{bold}{\encodingdefault}{\sfdefault}{bx}{n}

\usepackage{hyperref}
\usepackage{url}

\usepackage[utf8]{inputenc} 
\usepackage[T1]{fontenc}    
\usepackage{booktabs}       
\usepackage{amsfonts}       
\usepackage{nicefrac}       
\usepackage{microtype}      
\usepackage{xcolor}         
\usepackage{multirow}
\usepackage{graphicx}
\usepackage{subcaption}
\usepackage{amsmath}
\usepackage{wrapfig}
\usepackage{wrapfig}
\usepackage{enumitem}
\usepackage{algorithm}
\usepackage{algpseudocode} 
\usepackage{amssymb}
\usepackage{listings}
\usepackage[most]{tcolorbox}

\usepackage{colortbl}

\newcommand{\name}{\texttt{MemFit}}
\title{\name: Efficient Long-Term Agentic Memory}

\author{Mitchell Piehl \\
Department of Computer Science \\
The University of Iowa \\
\texttt{mpiehl@uiowa.edu}
\And
Muchao Ye \\
Department of Computer Science \\
The University of Iowa \\
\texttt{muchao-ye@uiowa.edu}
}

\iclrfinalcopy 
\begin{document}

\maketitle

\begin{abstract}
    Long-term memory systems for large language models (LLMs) have gained popularity for extending reasoning capabilities across applications. Current memory systems rely on LLM agents to organize and consolidate memory, resulting in costly, inefficient write operations. To address this limitation, we propose \name, a long-term memory system for conversational agents that reduces the cost and latency of memory operations. Unlike existing systems that rely on expensive LLM calls for memory construction or discard surface-level details through compression, \name{} stores each turn verbatim in an append-only store with near-instantaneous, LLM-free insertion, indexing turns with segment summaries rather than replacing them. Additionally, \name{} uses an LLM-free, multi-path retrieval strategy that combines lexical and semantic signals with cross-encoder reranking over caption-augmented episodes in both textual and multimodal settings. Empirical results on three widely used benchmarks, LoCoMo, MemGallery, and LongMemEval-S, show that \name{} achieves state-of-the-art performance while reducing memory construction time and cost several-fold, providing a scalable and efficient solution for persistent agentic memory.\footnote{The implementation code is publicly available in \url{https://github.com/mitchellpiehl/MemFit}}
\end{abstract}

\section{Introduction}
Large language models (LLMs) have demonstrated strong performance across diverse domains, but remain fundamentally constrained by fixed-length context windows, with performance degrading as token counts increase \citep{liu2024lost}. This degradation, often termed \emph{context rot} \citep{hong2025context}, reflects the model’s difficulty in reasoning over information buried in long prompts. To address this limitation, recent work equips LLMs with external long-term memory systems that store historical interactions and retrieve only the most relevant content at inference time \citep{packer2023memgpt, maharana2024evaluating, chhikara2025mem0, xu2025mem}, which has become a new trend in LLM research.
\begin{wrapfigure}{r}{0.6\textwidth}
\vspace{-0.1cm}
  \begin{center}
    \includegraphics[width=0.6\textwidth]{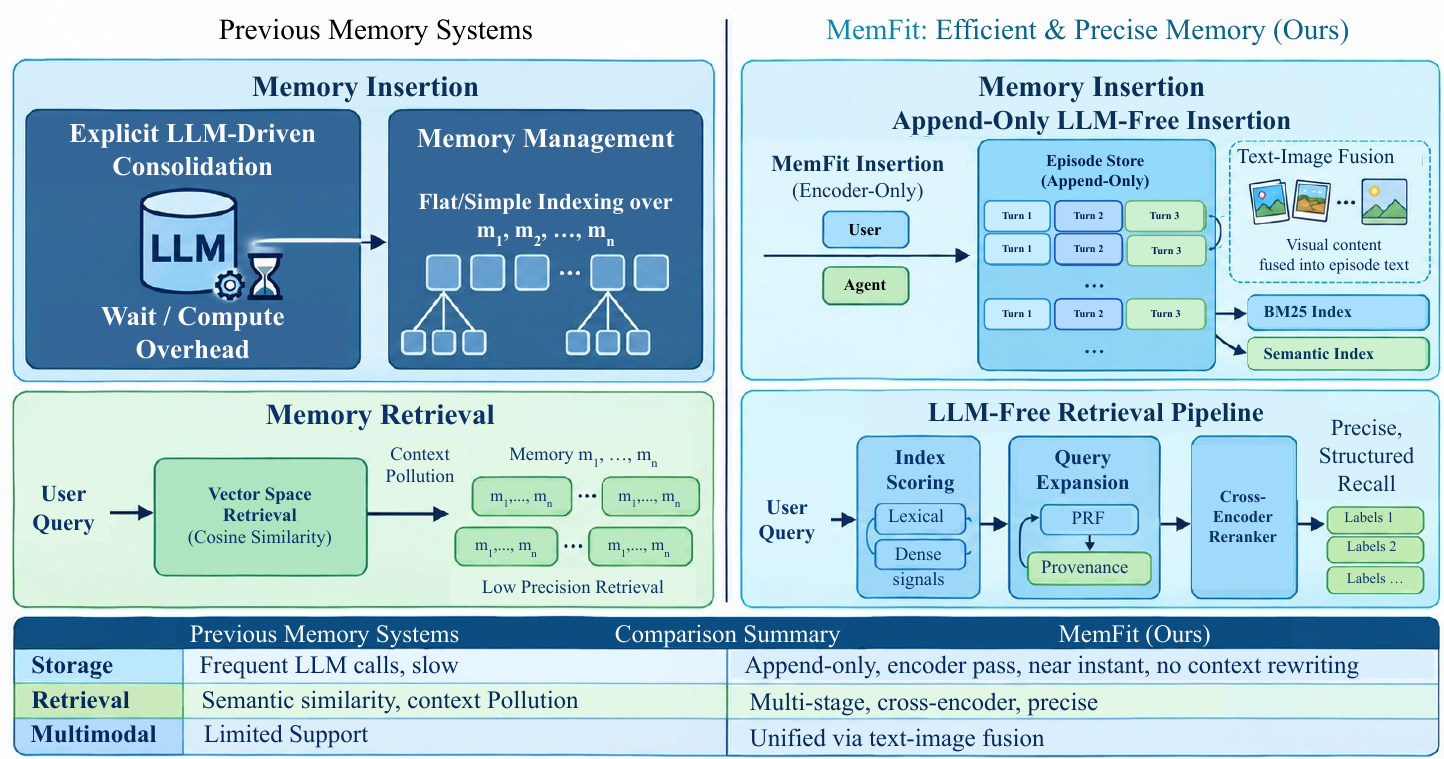}
  \end{center}
  \caption{Comparison between \name{} and existing memory system designs.}
  \label{fig:comparison}
  \vspace{-0.5cm}
\end{wrapfigure}

Existing memory architectures primarily differ in how they organize and maintain external memory. Systems such as MemGPT \citep{packer2023memgpt} and MemoryOS \citep{kang2025memory} adopt an operating-system perspective, delegating memory management decisions including what to retain, archive, or discard, to the LLM itself. In contrast, cognitively inspired approaches structure memory explicitly: LightMem \citep{fang2025lightmem} follows a staged design based on the Atkinson–Shiffrin model, while A-MEM \citep{xu2025mem} organizes information as an interconnected network of notes. Similarly, SimpleMem \citep{liu2026simplemem} focuses on improving storage efficiency via semantic lossless compression and multi-stage extraction. Despite these differences, most approaches rely on frequent LLM calls for memory consolidation and storage, and predominantly use semantic similarity for retrieval. This combination introduces two key limitations: high computational cost and susceptibility to context pollution, where irrelevant but semantically similar memories are retrieved. These limitations raise a central research question: 

\noindent \emph{Can a long-term memory system achieve efficient storage and precise retrieval without relying on frequent LLM calls?}

In this paper, we introduce \textbf{\name}, a long-term memory system for conversational agents that addresses this critical question by keeping the LLM out of the write path and out of the search itself. As illustrated in Figure \ref{fig:comparison}, firstly, our system replaces computationally expensive, LLM-driven storage with an \textbf{append-only episode store}: each turn is stored verbatim and indexed by a lexical signal, BM25 \citep{robertson2009bm25}, and an embedding signal, so insertion costs an encoder pass rather than an LLM call, and no stored content is rewritten as the conversation grows. A second layer summarizes each conversational segment in a single call and records the turns it was built from. This means that the summaries index the store rather than compress it, leaving the original interaction content retrievable. Since nothing is merged or discarded at write time, a superseded fact and its correction both survive with their dates, and the choice between them is deferred until the question is known. Secondly, to ensure precise retrieval, \name{} uses an \textbf{LLM-free retrieval pipeline}, in which both indexes score the store, two additive expansions widen the candidate pool, and a cross-encoder \citep{nogueira2019passage} decides the final order. The system further supports both textual and multimodal settings via a \emph{text-image fusion} design, where detailed image captions are embedded directly into the searchable episode text, enabling visual content to be retrieved through the same lexical and semantic pathways as text. 

To summarize, the primary contributions of this work are as follows:
\begin{itemize}[leftmargin=*]
    \item We introduce \name{}, an efficient and persistent memory architecture that significantly reduces computational overhead while improving retrieval accuracy.
    \item The proposed \name{} is a versatile memory system that extends beyond textual reasoning to support complex multimodal tasks within a unified framework, representing one of the few agentic memory systems designed for multimodal retrieval and reasoning.
    \item Lastly, \name{} achieves a superior balance of retrieval efficiency and reasoning correctness compared to memory systems of comparable cost on representative long-term conversational benchmarks including LoCoMo \citep{maharana2024evaluating}, MemGallery \citep{bei2026mem}, and LongMemEval-S \citep{wu2024longmemeval} datasets, under one configuration across every backbone we evaluate, spanning three model families.
\end{itemize}

\section{Related work}

\noindent \textbf{Memory systems for LLMs}. Memory systems for LLM agents are closely related to Retrieval-Augmented Generation (RAG) \citep{gao2023retrieval, lin2023ra}, as both rely on retrieving relevant external context to ground model responses. Foundational RAG work has shown the effectiveness of augmenting language models with external textual corpora \citep{lewis2020retrieval}, a paradigm that is now adopted in agentic memory systems. Early approaches evolved from simple sliding-window contexts to more structured, persistent memory architectures. For example, A-MEM~\citep{xu2025mem} organizes memory as a graph-based Zettelkasten with dynamically generated links to support memory evolution, while Mem0~\citep{chhikara2025mem0} introduces an incremental pipeline for extraction and consolidation across sessions. Other work focuses on efficiency. For example,  SimpleMem~\citep{liu2026simplemem} applies semantic lossless compression to reduce token overhead, and LightMem~\citep{fang2025lightmem} decouples memory maintenance from inference via a "sleep-time" consolidation process. In contrast to these approaches, \name{} never rewrites what it stores. Instead, it keeps turns verbatim without per-turn LLM calls, and its segment summaries index the stored turns.

While RAG has recently expanded beyond text to support multimodal retrieval (e.g., MuRAG~\citep{chen2022murag}, RA-CM3~\citep{yasunaga2022retrieval}, and UniversalRAG~\citep{yeo2025universalrag}), this capability remains underexplored in agentic memory systems. 
\name{} bridges this gap through a \emph{text-image fusion} design that embeds detailed visual descriptions directly into the episode text. This enables fine-grained visual retrieval within a unified text-based search space, avoiding the limitations of standard cross-modal embedding approaches.

\noindent \textbf{Memory summarization and consolidation}.
Researchers have explored summarizing or restructuring stored content with an LLM to manage large conversational contexts. For example, RAPTOR \citep{sarthi2024raptor} builds a tree by recursively clustering and summarizing text segments, allowing for retrieval at varying levels of abstraction. Similarly, MemTree \citep{rezazadeh2024isolated}, H-MEM \citep{sun2025hierarchical}, and HiMem \citep{zhang2026himem} organize conversations into hierarchies whose upper levels an LLM writes, or rewrites, as memories arrive. Consolidation goes further by altering stored content itself: MemoryBank \citep{zhong2024memorybank} lets memories fade along a forgetting curve, and Mem0 deletes memories contradicted by new information. A close precedent to our design is the memory stream of Generative Agents \citep{park2023generative}, which keeps a complete record of experiences and synthesizes higher-level reflections over it by rating each memory's importance with an LLM as it is recorded. To increase efficiency, \name{} merges and deletes nothing, and segments the conversation without an LLM by applying TextTiling \citep{hearst1997text} to episode embeddings. Unlike SeCom \citep{pan2025secom}, which retrieves compressed segments as memory units, each segment summary indexes the verbatim turns it describes.

\noindent\textbf{Iterative retrieval for LLMs}.
Iterative retrieval methods refine search through repeated LLM feedback. Self-RAG~\citep{asai2023self} employs self-reflection to improve retrieval quality, FLARE~\citep{jiang2023active} generates intermediate predictions to guide subsequent queries, and IRCoT~\citep{trivedi2023interleaving} interleaves retrieval with chain-of-thought steps for multi-hop questions. MemWalker~\citep{chen2023walking} navigates a tree of summaries through step-by-step LLM reading, and Recursive Language Models (RLMs)~\citep{zhang2025recursive} frame long-context reasoning as iterative exploration over external inputs. However, these approaches rely on multiple sequential LLM invocations, which increases latency and cost. In contrast, \name{} keeps the LLM out of the search itself, relying on a normalized convex fusion of lexical and dense scores \citep{bruch2023analysis}, pseudo-relevance feedback for multi-hop questions, and a cross-encoder reranker \citep{nogueira2019passage}, and achieves sophisticated retrieval without LLM guidance.

\begin{figure}
  \centering \includegraphics[width=\textwidth]{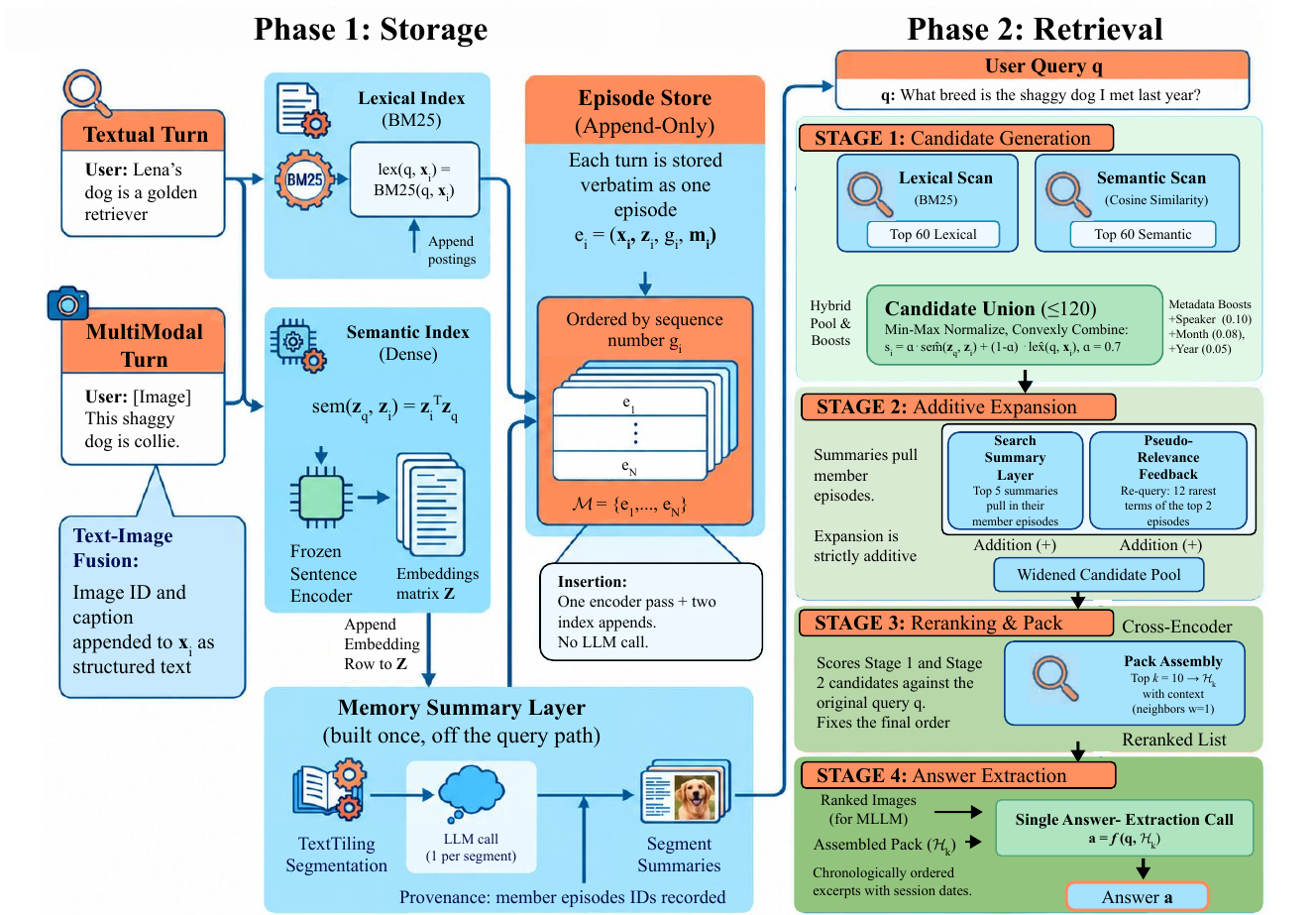}
  \caption{Overview of \name. Storage is LLM-free apart from one summary call per conversational segment: each turn is stored verbatim as an episode and indexed by a lexical and a semantic signal. Retrieval scores the stored episodes on both signals, widens the candidate pool through summary provenance and pseudo-relevance feedback, and reranks with a cross-encoder.}
  \label{fig:main}
\end{figure}

\section{Methodology}

We present \name, a long-term memory system for conversational agents that utilizes a write path that never calls an LLM per turn with single-call retrieval. The system can flexibly operate in two modes: a base \emph{textual} mode for purely textual conversations, and a \emph{multimodal} mode that additionally handles shared images. Both modes share the same storage, indexing, and retrieval procedure, while the multimodal mode extends each stage with image captions and, at answer time, the images themselves.

\subsection{Problem statement}

A conversational memory system operates over a growing dialogue history $\mathcal{H} = \{(u_i, t_i, p_i, v_i)\}_{i=1}^{N}$ of utterances $u_i$ with timestamps $t_i$, speakers $p_i$, and optional images $v_i$, and must support two operations: \emph{storage}, which organizes $\mathcal{H}$ into a persistent structure $\mathcal{M}$ that grows incrementally, and \emph{retrieval}, which given a query $q$ returns an answer $a$ grounded in $\mathcal{M}$. This process can be viewed as selecting a small subset of relevant episodes $\mathcal{H}_k$ from $\mathcal{H}$ and generating an answer via  $a = f(q, \mathcal{H}_k)$,
where $f$ denotes the underlying LLM conditioned on both the query and the retrieved memory. Queries range from direct factual questions (``What is the name of Lena's dog?''), temporal reasoning (``When did Lena adopt her dog?''), multi-hop synthesis (``Which two dogs are described as intelligent?''), to multimodal tasks that require connecting a new image to previously discussed visual content (``What breed is the dog in this picture?''). To answer such queries, a memory system must \emph{(1) locate the relevant episodes from potentially thousands of stored interactions}, \emph{(2) reconcile information that may have been updated across sessions}, and \emph{(3) connect visual content to its surrounding conversational context in multimodal settings}. Success requires both \textbf{inexpensive storage}, so that every turn can be absorbed as it arrives without an LLM call of its own, and \textbf{precise retrieval}, so that the correct episode is surfaced among many candidates. Specifically, \name{} achieves them via an \textbf{append-only episode store} carrying one sparse and one dense index, a layer of \textbf{segment summaries} that serves as an index into that store rather than as a replacement for it (detailed in \S~\ref{sec:storage}), and a \textbf{single answer-extraction call} that reads the retrieved excerpts (detailed in \S~\ref{sec:retrieval}).

\subsection{Memory storage and the summary layer}
\label{sec:storage}
\noindent \textbf{Memory representation}.
Each utterance in $\mathcal{H}$ becomes one \emph{memory episode} written as $e_i = (\mathbf{x}_i, \mathbf{z}_i, g_i, \mathbf{m}_i)$, where the formatted text $\mathbf{x}_i$ is labeled with time and speaker (e.g., \emph{[Time: 2023-05-12] [Speaker: Lena] I adopted a golden retriever yesterday}.), $\mathbf{z}_i = f_{\text{enc}}(\mathbf{x}_i) \in \mathbb{R}^d$ is the $L_2$-normalized embedding of $\mathbf{x}_i$ from a frozen sentence encoder $f_{\text{enc}}$, $g_i$ is a global sequence number for temporal ordering, and $\mathbf{m}_i$ is the structured metadata, including fields such as timestamps, speaker identifiers, and session and turn identifiers used for filtering and scoring. This granularity is deliberate because it matches the unit each benchmark annotates as gold evidence and the unit an answer usually needs, so no answer is split across a boundary the system chose for itself, and retrieval is scored at the granularity the benchmark defines. In the multimodal setting, an image-bearing turn additionally carries its image identifier and caption, appended to $\mathbf{x}_i$ as structured text (\texttt{[Image D1:IMG\_001: a golden retriever on a beach\dots]}), which makes image content searchable through the same lexical and semantic pathways used for text; we term this \emph{text-image fusion}. We source captions from ground-truth annotations when available; otherwise, we generate them with a multimodal large language model (MLLM).

\noindent \textbf{Episode store and indexes}.
Episodes are stored in an \emph{append-only store} $\mathcal{M} = \{e_1, \dots, e_N\}$ under two indexes that answer different kinds of questions about the same text. The first is a BM25 index over the tokens of $\mathbf{x}_i$, holding the term frequencies of each episode and the document frequency of each term, so that a bag of query terms scores the store in one pass over the postings of those terms, touching only the episodes in which one of them occurs; this is the signal that matches a name, a date, or a rare noun that has to appear verbatim. The second is a dense index: the $L_2$-normalized embeddings $\mathbf{z}_i$ stacked into a matrix $\mathbf{Z} \in \mathbb{R}^{N \times d}$ over the $N$ stored episodes, so that scoring the store against a query embedding is the single matrix-vector product $\mathbf{Z}\mathbf{z}_q$, whose $i$-th entry is the cosine similarity of episode $i$. This signal matches a paraphrase that shares no words with the question. Inserting an episode adds one row to $\mathbf{Z}$ and its postings to the index, and nothing already stored is rewritten, re-embedded, or re-summarized when later turns arrive, so the memory absorbs a conversation at the rate it is produced. The sequence numbers $g_i$ preserve chronological order, which retrieval uses to return a retrieved turn together with its immediate neighbors.

\noindent \textbf{Segment summaries}.
A second memory layer describes stretches of conversation, and is the only place an LLM touches the write path. TextTiling \citep{hearst1997text} segments consecutive episodes without an LLM by comparing their embeddings across each gap. At every gap, the cosine similarity between the mean embeddings of the $b$ episodes on either side gives a coherence score, with $b = \max(3, \mathrm{round}(\sqrt{n}/2))$ for $N$ stored episodes, and a boundary is placed at each local minimum whose depth below its neighboring peaks exceeds $\mu_d + \sigma_d/2$, where $\mu_d$ and $\sigma_d$ are the mean and standard deviation of the depth over all gaps. Every resulting segment of at least three episodes is then summarized in a single LLM call that sees up to $14$ evenly spaced member excerpts and returns two sentences ($\leq 200$ tokens). Each summary is stored alongside the episodes, with its members' identifiers as provenance and its own embedding, and retrieved by the same hybrid score. Summaries deliberately do not compress the store: the episodes they describe remain verbatim and separately retrievable, because the summary \emph{finds} a stretch of conversation while the raw turn \emph{answers} the question.

\noindent \textbf{Updates without consolidation}.
The memory store intentionally has no merge, rewrite, or delete operation. When a fact changes, the store keeps the original statement and its correction as separate episodes with their own dates, and no contradiction must be detected at write time to keep the store consistent. What stands in for consolidation is the shape of what retrieval returns: competing statements about one subject tend to share the content words a query about that subject uses, so the score that surfaces one tends to surface the other, and the retrieved excerpts are ordered chronologically with each excerpt's session date attached, leaving the reader the old value, the new value, and the order between them at the moment the question is known. A consolidator chooses which statement survives before it knows what will be asked, and the evidence it discards is unrecoverable; deferring that choice costs nothing, because retrieval already returns both sides. Summaries serve the purpose of consolidation, acting as a compact description of a stretch of conversation while keeping the property recall depends on, since they index the turns they describe rather than standing in for them.

\subsection{Memory retrieval for answer extraction}
\label{sec:retrieval}
Once the store has been built, it is used for all downstream query answering. When a user or evaluation benchmark poses a query $q$ (e.g., a factual question, a temporal reasoning task, or a visual search over shared images), retrieval proceeds in four stages, as illustrated in Figure~\ref{fig:main}: candidate generation over both indexes, two additive expansions of the candidate pool, a cross-encoder rerank that fixes what the reader sees, and then final answer extraction.

\noindent \textbf{Stage 1: Hybrid candidate generation}. Let $\mathbf{z}_q = f_{\text{enc}}(q)$ denote the $L_2$-normalized query embedding produced by the same frozen sentence encoder used during storage. Two paths score the stored episodes: a \emph{lexical path}, $\text{lex}(q, \mathbf{x}_i) = \text{BM25}(q, \mathbf{x}_i)$, and a \emph{semantic path}, the cosine similarity $\text{sem}(\mathbf{z}_q, \mathbf{z}_i) = \mathbf{z}_i^{\top}\mathbf{z}_q$, computed for every episode at once as the matrix-vector product $\mathbf{Z}\mathbf{z}_q$. The top $n_{\text{scan}} = 60$ episodes from each path form a candidate union, over which the two scores are min-max normalized and combined convexly,
\begin{equation}
\label{eq:fusion}
    s_i = \alpha \cdot \widehat{\text{sem}}(\mathbf{z}_q, \mathbf{z}_i) \; + \; (1-\alpha) \cdot \widehat{\text{lex}}(q, \mathbf{x}_i),
\end{equation}
where $\widehat{\text{sem}}$ and $\widehat{\text{lex}}$ denote the two scores after min-max normalization over the union, and $\alpha$ is a hyperparameter  (by default $\alpha=0.7$). Since BM25 is unbounded while cosine similarity is not, their raw values share no scale, so normalizing before combining makes the two signals comparable. Each candidate additionally receives additive metadata boosts when the query mentions a specific speaker ($+0.10$), year ($+0.05$), or month ($+0.08$) that matches the episode's metadata, encouraging episodes from the queried time period or speaker to surface. Because the boosts are applied after normalization, a boost is a fixed fraction of the observed score range rather than an arbitrary offset between two scales.

\noindent \textbf{Stage 2: Additive candidate expansion}. In practice, candidate generation is the binding constraint for retrieval: far more questions are lost because the gold turn never enters the candidate pool than because it is ranked too low within it. To address this, two expansions widen the pool. First, the summary layer is searched with the same hybrid score, and the highest-scoring summaries pull their member episodes into the pool through their provenance. This means that a summary that matches the query retrieves the turns it was built from. Second, pseudo-relevance feedback re-runs candidate generation with the $12$ rarest terms, by inverse document frequency, from the two top-ranked episodes; this targets multi-hop questions, whose two supporting turns lie in different sessions where a single query finds only one. Both expansions are strictly \emph{additive} because they append results to the candidate pool. Therefore, expansion can only add evidence the first stage missed and can never evict what it found.

\noindent \textbf{Stage 3: Reranking and pack assembly}. Once the candidate pool is generated, a cross-encoder scores the top $50$ first-stage candidates, together with every expansion candidate, against the \emph{original} question. The lexical and semantic scores therefore decide only membership in the candidate pool, not the final order, so their differing scales stop affecting the ranking. Because a cross-encoder scores a two-sentence summary and a raw turn on different text distributions, we standardize scores within each kind before ordering, so a summary holds a pack slot only if it ranks as well among summaries as an episode does among episodes. The $k = 10$ highest-ranked candidates form the collection $\mathcal{H}_k$, and every retrieved episode brings the turns within $w=1$ of it in the same session, restoring the context a short reply depends on without a second search.

\noindent \textbf{Stage 4: Answer extraction}. With the top $k$ highest-ranked candidates, the collection is rendered as chronologically ordered excerpts, each stamped with its session date and with relative time expressions pre-resolved to absolute dates, and the reader extracts the answer $a = f(q, \mathcal{H}_k)$. One prompt serves every question. In multimodal mode, the reader is an MLLM, and the images of the ranked episodes, together with any image attached to the query, are passed in that same call, each labeled with its image identifier; captions remain in the episode text, so retrieval is unchanged.

\noindent \textbf{Discussion on efficiency}. Neither paths calls an LLM per turn or per search step. Insertion is an embedding and two index updates, and no stored episode is ever revisited. The only write-time LLM cost is one call per segment. Retrieval relies on arithmetic to retrieve memories, and only the sparse scan and the matrix-vector product grow with memory size, both of which grow linearly. 

\section{Experiments and results}

\subsection{Experimental settings}

\paragraph{Benchmark datasets.} We evaluate the performance of \name{} on three diverse benchmarks designed to stress-test long-term  memory as follows. More details are shown in Appendix \S~\ref{sec:dataset}.
\begin{itemize}[leftmargin=*]
    \item \textbf{LoCoMo} \citep{maharana2024evaluating}: A long-term conversational memory benchmark consisting of multi-session dialogues. It evaluates five question types: single-hop, multi-hop, temporal reasoning, open-domain knowledge, and adversarial. Consistent with prior work ~\citep{chhikara2025mem0, liu2026simplemem}, we exclude adversarial questions.
    \item \textbf{MemGallery} \citep{bei2026mem}: A multimodal benchmark for evaluating long-term memory in MLLM agents that evaluates three functional dimensions: memory extraction, multimodal reasoning, and dynamic knowledge management.
    \item \textbf{LongMemEval-S} \citep{wu2024longmemeval}: A benchmark of 500 questions, each asked over its own user–assistant chat history of about 50 sessions and 115k tokens, covering six question types: single-session user, assistant, and preference; multi-session; knowledge update; and temporal reasoning.
\end{itemize}

\paragraph{Baselines.} We compare \name{} against  leading memory systems, MemoryBank \citep{zhong2024memorybank}, MemGPT \citep{packer2023memgpt}, MemoryOS \citep{kang2025memory}, A-MEM \citep{xu2025mem}, LightMem \citep{fang2025lightmem}, Mem0 \citep{chhikara2025mem0}, and SimpleMem \citep{liu2026simplemem}, and against full-context and naive retrieval (NaiveRAG) baselines where a benchmark reports them. Details on each are provided in Appendix \S~\ref{sec:baseline}.

\paragraph{Backbones and evaluation metrics.} To show that \name{} is backend-agnostic, we evaluate it with the \textbf{proprietary} \texttt{GPT-4o}, \texttt{GPT-4.1-mini}, and \texttt{GPT-4o-mini} \citep{achiam2023gpt}, the \textbf{open-source} \texttt{Qwen3-8B} \citep{yang2025qwen3} and \texttt{Gemma3-27B} \citep{team2025gemma}, and the \textbf{multimodal} \texttt{Qwen2.5-VL-7B} \citep{baiQwen25} on MemGallery. On LoCoMo, we report F1 and BLEU-1 \citep{papineni2002bleu} per category and averaged over the four categories; on MemGallery, token-level F1 over all questions, as defined by the benchmark; and on LongMemEval-S, accuracy graded by an LLM judge with the benchmark's official prompts, with \texttt{GPT-4o-mini} as both backbone and judge to match the published baselines. Ablations also report LLM-judge accuracy (Appendix~\ref{sec:prompt}). For efficiency, we measure LLM calls, input and output tokens, cost, memory build time, and per-question latency.

\begin{table}[t]
\centering
\caption{Performance on the LoCoMo benchmark with four backbones. F1 and BLEU-1 per category and averaged over the four categories. {\name} achieves superior performance. Best result per backbone in bold.}
\resizebox{\textwidth}{!}{
\renewcommand{\arraystretch}{1.15}
\begin{tabular}{ll|cccccccc|cc}
\toprule
\multirow{2}{*}{Model} & \multirow{2}{*}{Method} & \multicolumn{2}{c}{Multi-Hop} & \multicolumn{2}{c}{Temporal} & \multicolumn{2}{c}{OpenDomain} & \multicolumn{2}{c|}{Single-Hop} & \multicolumn{2}{c}{Average} \\
\cmidrule(lr){3-4} \cmidrule(lr){5-6} \cmidrule(lr){7-8} \cmidrule(lr){9-10} \cmidrule(lr){11-12}
& & F1 (\%) & BLEU (\%) & F1 (\%) & BLEU (\%) & F1  (\%) & BLEU (\%)  & F1 (\%) & BLEU (\%) & F1 (\%) & BLEU (\%) \\
\midrule
\multirow{7}{*}{\rotatebox{90}{GPT-4.1-mini}} 
& MemoryBank & 5.00 & 4.68 & 5.94 & 4.78 & 5.16 & 4.52 & 5.72 & 4.86 & 5.46 & 4.71 \\
& MemGPT & 17.72 & 16.02 & 19.44 & 16.54 & 11.29 & 10.18 & 25.59 & 24.25 & 18.51 & 16.75 \\
& A-MEM & 25.06 & 17.32 & 51.01 & 44.75 & 13.22 & 14.75 & 41.02 & 36.99 & 32.58 & 28.45 \\
& LightMem & 24.96 & 21.66 & 20.55 & 18.39 & 19.21 & 17.68 & 33.79 & 29.66 & 24.63 & 21.85 \\
& Mem0 & 30.14 & 27.62 & 48.91 & 44.82 & 16.43 & 14.94 & 41.3 & 36.17 & 34.20 & 30.89 \\
& SimpleMem & \textbf{43.46} & \textbf{38.82} & 58.62 & 50.10 & 19.76 & 18.04 & 51.12 & 43.53 & 43.24 & 37.62 \\
\rowcolor{cyan!20}
& \textbf{\name} & 41.58 & 33.70 & \textbf{66.05} & \textbf{60.99} & \textbf{28.85} & \textbf{25.42} & \textbf{64.56} & \textbf{58.84} & \textbf{50.26} & \textbf{44.74} \\
\midrule
\multirow{7}{*}{\rotatebox{90}{GPT-4o}}
& MemoryBank & 6.49 & 4.69 & 2.47 & 2.43 & 6.43 & 5.30 & 8.28 & 7.10 & 5.92 & 4.88 \\
& MemGPT & 30.36 & 22.83 & 17.29 & 13.18 & 12.24 & 11.87 & 40.16 & 36.35 & 25.01 & 21.06 \\
& A-MEM & 32.86 & 23.76 & 39.41 & 31.23 & 17.10 & 15.84 & 44.43 & 38.97 & 33.45 & 27.45 \\
& LightMem & 28.15 & 21.83 & 36.53 & 29.12 & 13.38 & 11.54 & 33.76 & 28.02 & 27.96 & 22.63 \\
& Mem0 & 35.13 & 27.56 & 52.38 & 44.15 & 17.73 & 15.92 & 39.12 & 35.43 & 36.09 & 30.77 \\
& SimpleMem & 35.89 & 32.83 & 56.71 & 20.57 & 18.23 & 16.34 & 45.41 & 39.25 & 39.06 & 27.25 \\
\rowcolor{cyan!20}
& \textbf{\name} & \textbf{41.28} & \textbf{33.07} & \textbf{65.48} & \textbf{60.21} & \textbf{25.38} & \textbf{22.33} & \textbf{65.16} & \textbf{60.23} & \textbf{49.33} & \textbf{43.96} \\
\midrule
\multirow{7}{*}{\rotatebox{90}{Qwen3:8b}}
& MemoryBank & 9.50 & 7.10 & 3.80 & 2.50 & 7.50 & 6.50 & 9.20 & 7.50 & 7.50 & 5.90 \\
& MemGPT & 14.20 & 9.80 & 5.50 & 4.20 & 12.50 & 10.80 & 11.50 & 9.10 & 10.93 & 8.48 \\
& A-MEM & 20.50 & 13.80 & 22.50 & 18.20 & 13.20 & 10.50 & 26.80 & 21.50 & 20.75 & 16.00 \\
& LightMem & 18.53 & 14.23 & 26.78 & 21.52 & 14.12 & 11.24 & 29.48 & 23.83 & 22.23 & 17.71 \\
& Mem0 & 22.42 & 16.83 & 32.48 & 26.13 & 15.23 & 12.54 & 33.05 & 27.24 & 25.80 & 20.69 \\
& SimpleMem & 28.97 & 24.93 & 42.85 & 36.49 & 15.35 & 13.90 & 46.62 & 40.69 & 33.45 & 29.00 \\
\rowcolor{cyan!20}
& \textbf{\name} & \textbf{40.22} & \textbf{33.49} & \textbf{59.26} & \textbf{54.02} & \textbf{16.66} & \textbf{14.48} & \textbf{60.63} & \textbf{55.51} & \textbf{44.19} & \textbf{39.38} \\
\midrule
\multirow{4}{*}{\rotatebox{90}{\footnotesize \shortstack{Gemma3 \\ 27b}}}
& A-MEM & 22.29 & 12.83 & 26.40 & 20.95 & 7.97 & 6.45 & 30.80 & 26.28 & 21.87 & 16.63 \\
& Mem0 & 25.32 & 18.89 & 23.99 & 17.50 & 12.22 & 10.30 & 30.84 & 27.42 & 23.09 & 18.53 \\
& SimpleMem & 28.35 & 23.46 & 39.14 & 28.49 & 10.41 & 7.65 & 38.00 & 31.32 & 28.98 & 22.73 \\
\rowcolor{cyan!20}
& \textbf{\name} & \textbf{40.45} & \textbf{33.27} & \textbf{58.93} & \textbf{53.18} & \textbf{18.42} & \textbf{15.14} & \textbf{63.88} & \textbf{58.11} & \textbf{45.42} & \textbf{39.92} \\
\bottomrule
\end{tabular}
}
\label{tab:locomo1}
\end{table}

\paragraph{Implementation details.} We use frozen \texttt{all-MiniLM-L6-v2} (384-dimensional) for episode, query, and summary embeddings, and the frozen cross-encoder \texttt{ms-marco-MiniLM-L-6-v2} for reranking. Candidate generation keeps the top $n_{\text{scan}}{=}60$ episodes from each of the lexical and semantic paths and fuses them with $\alpha{=}0.7$; the cross-encoder scores the top $50$ of the pool, in which the $10$ best-matching summaries are interleaved with the episodes, together with every expansion candidate; the pack is $k{=}10$ episodes with a context window of $w{=}1$. Pseudo-relevance feedback uses the $12$ rarest terms of the top $2$ episodes, and the $5$ best-matching summaries expand the pool through their provenance. Then, segments that have at least three episodes are summarized from at most $14$ excerpts into at most $200$ tokens.

We use one configuration for every backbone and all benchmarks, with nothing tuned per corpus. For open-source models, we use Ollama as the backbone. All answer extraction uses greedy decoding; we pre-resolve relative time expressions in retrieved excerpts to absolute dates in Python before the extraction call, and we pair $p$-values with sign-flip permutation tests over the per-question scores.

\subsection{Comparison results}

When comparing \name{} to the various leading memory systems, \name{} achieves the best overall performance on all three benchmarks and with every backbone, as shown in Table~\ref{tab:locomo1} and Table~\ref{tab:overall}.

\paragraph{Comparison results on LoCoMo.}
As shown in Table~\ref{tab:locomo1}, \name{} achieves state-of-the-art performance on the LoCoMo benchmark across diverse backbone models. Using \texttt{Qwen3-8B}, \name{} yields an average F1 score of 44.19\%, a significant improvement over the second-best baseline SimpleMem (33.45\%), and the same ordering holds for \texttt{GPT-4.1-mini} (50.26\% vs.\ 43.24\%), \texttt{GPT-4o} (49.33\% vs.\ 39.06\%), and \texttt{Gemma3-27B} (45.42\% vs.\ 28.98\%). The margin is widest on the open-source backbones because \name{} degrades far less when the reader is swapped. For example, moving from \texttt{GPT-4.1-mini} to \texttt{Gemma3-27B} costs \name{} 4.84 average F1 and costs the strongest baseline 14.26. Per category, \name{} leads on Temporal, OpenDomain, and Single-Hop queries with every backbone. Multi-hop is the one category \name{} does not lead in every setting, 
\begin{wraptable}{r}{0.46\textwidth}
\centering
\caption{Overall results on MemGallery~\citep{bei2026mem} and
LongMemEval-S~\citep{wu2024longmemeval}.}
\label{tab:overall}
\small
\setlength{\tabcolsep}{8pt}
\renewcommand{\arraystretch}{1.05}

\textbf{(a) MemGallery (\texttt{Qwen2.5-VL-7B})}

\vspace{2pt}
\begin{tabular}{p{2.5cm}>{\centering\arraybackslash}p{2.0cm}}
\toprule
Method & F1 (\%) $\uparrow$ \\
\midrule
Full context & 36.25 \\
NaiveRAG & 59.74 \\
\addlinespace[2pt]
MemGPT & 52.82 \\
A-MEM & 62.28 \\
MemoryOS & 61.09 \\
\midrule
\cellcolor{cyan!20}\textbf{\name} &
\cellcolor{cyan!20}\textbf{67.66} \\
\bottomrule
\end{tabular}

\vspace{6pt}

\textbf{(b) LongMemEval-S (\texttt{GPT-4o-mini})}

\vspace{2pt}
\begin{tabular}{p{2.5cm}>{\centering\arraybackslash}p{2.0cm}}
\toprule
Method & Accuracy (\%) $\uparrow$ \\
\midrule
Full context & 56.80 \\
NaiveRAG & 61.00 \\
\addlinespace[2pt]
A-MEM & 62.60 \\
LightMem & 64.29 \\
Mem0 & 53.61 \\
MemoryOS & 44.80 \\
LangMem & 37.20 \\
\midrule
\cellcolor{cyan!20}\textbf{\name} &
\cellcolor{cyan!20}\textbf{64.60} \\
\bottomrule
\end{tabular}

\vspace{-35pt}
\end{wraptable}
trailing SimpleMem with \texttt{GPT-4.1-mini} (41.58\% vs.\ 43.46\%): such questions require synthesizing facts across separately stored episodes, and methods that consolidate memories through LLM calls during storage can pre-compute those connections, whereas \name{} composes at retrieval time through pseudo-relevance feedback.

\noindent \textbf{Comparison results on MemGallery and LongMemEval-S}.
On the MemGallery benchmark (Table~\ref{tab:overall}), \name{} achieves the highest overall F1 among the compared memory systems (67.66\%), ahead of A-MEM (62.28\%) and MemoryOS (61.09\%), and on LongMemEval-S \citep{wu2024longmemeval}, where every question comes with its own history of roughly 115k tokens, \name{} with \texttt{GPT-4o-mini} answers 64.60\% of the 500 questions correctly, beating every baseline. \name{} is 0.31 points above LightMem, 2.0 points above A-MEM, 3.6 points above naive retrieval, 7.8 points above reading the full history, and 11.0 points above Mem0. 

\begin{figure*}[t]
    \centering
    \begin{minipage}[b]{0.19\textwidth}
        \centering
        \begin{subfigure}[b]{\textwidth}
            \centering
            \includegraphics[width=\textwidth]{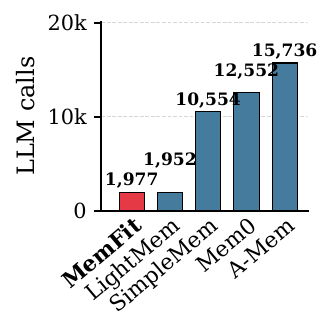}
            \caption{LLM Calls}
            \label{fig:llm_calls}
        \end{subfigure}
        \vspace{1em}
        \begin{subfigure}[b]{\textwidth}
            \centering
            \includegraphics[width=\textwidth]{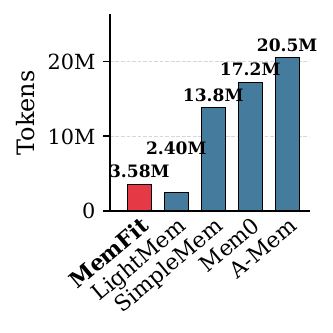}
            \caption{Input Tokens}
            \label{fig:input_tokens}
        \end{subfigure}
    \end{minipage}
    \hfill
    \begin{minipage}[b]{0.19\textwidth}
        \centering
        \begin{subfigure}[b]{\textwidth}
            \centering
            \includegraphics[width=\textwidth]{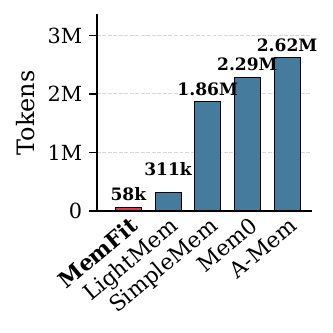}
            \caption{Output Tokens}
            \label{fig:output_tokens}
        \end{subfigure}
        \vspace{1em}
        \begin{subfigure}[b]{\textwidth}
            \centering
            \includegraphics[width=\textwidth]{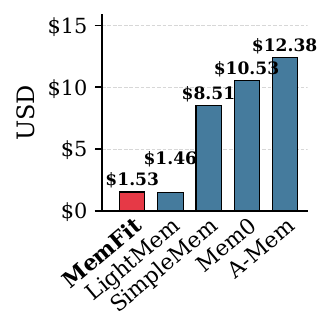}
            \caption{Cost (USD)}
            \label{fig:cost}
        \end{subfigure}
    \end{minipage}
    \hfill
    \begin{minipage}[b]{0.60\textwidth}
        \centering
        \begin{subfigure}[b]{\textwidth}
            \centering
            \includegraphics[width=\textwidth]{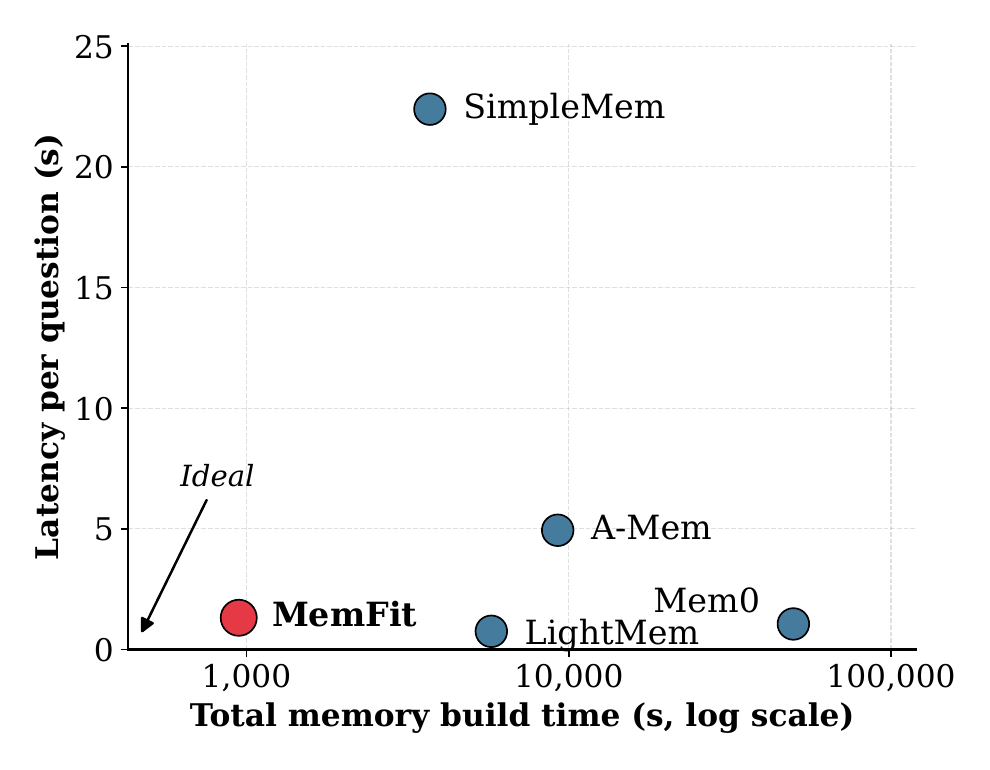}
            \caption{Latency Trade-off: Storage vs. Retrieval}
            \label{fig:latency_tradeoff}
        \end{subfigure}
    \end{minipage}

    \caption{Efficiency on LoCoMo with \texttt{Gemma3-27B}, covering memory construction for all ten conversations and all questions. (a--c) LLM calls and tokens. (d) Cost at \texttt{GPT-4.1-mini} list prices (\$0.40 and \$1.60 per million input and output tokens). (e) Memory build time against latency.}
    \label{fig:efficiency_and_latency}
\end{figure*}

\subsection{Computational efficiency analysis}

In Fig.~\ref{fig:efficiency_and_latency}, we compare the efficiency of \name{} with \texttt{SimpleMem} \citep{liu2026simplemem} and \texttt{LightMem} \citep{fang2025lightmem}, both designed for efficiency, \texttt{A-MEM} \citep{xu2025mem}, and \texttt{Mem0} \citep{chhikara2025mem0} on LoCoMo with \texttt{Gemma3-27B}, counting memory construction for the entire dataset.

\paragraph{Resource consumption.}
Against the three systems that consult an LLM as memories arrive, \name{} issues 81--87\% fewer LLM calls (1,977 vs.\ 10,554--15,736; Fig.~\ref{fig:llm_calls}), generates $32$--$45\times$ fewer output tokens, and costs 82--88\% less at \texttt{GPT-4.1-mini} prices (\$1.53 vs.\ \$8.51--\$12.38; Fig.~\ref{fig:cost}). \texttt{LightMem} has a comparable budget: both systems issue about 400 write-time calls and one call per question, but \name{}'s write path uses $5.1\times$ fewer tokens (0.22M vs.\ 1.13M) and costs $5.7\times$ less (\$0.14 vs.\ \$0.81), because its only write-time output is one summary per segment, while \texttt{LightMem} rewrites the conversation into memory entries. In exchange, \name{}'s reader receives verbatim turns rather than distilled facts (2,210 vs.\ 1,015 input tokens per question), so full runs cost about the same (\$1.53 vs.\ \$1.46): \name{} pays less to build memory and more to read it.

\paragraph{Latency.}
\name{} builds memory $3.9$--$52.6\times$ faster than every baseline (Fig.~\ref{fig:latency_tradeoff}): 946s for the whole benchmark, of which LLM-free storage and indexing take 15s, against 3,708s for \texttt{SimpleMem}, 5,750s for \texttt{LightMem}, 9,240s for \texttt{A-MEM}, and 49,765s for \texttt{Mem0}. Per question, it answers in 1.23s, faster than \texttt{A-MEM} (4.94s) and \texttt{SimpleMem} (22.39s) but slower than \texttt{Mem0} (1.06s) and \texttt{LightMem} (0.75s), trading a somewhat longer read for the fastest write path.

\begin{table}[t]
\centering
\caption{\textbf{Left:} Ablation of \name{} on LoCoMo with \texttt{Gemma3-27B}; removing any component lowers judge accuracy, F1, and BLEU. \textbf{Right:} Sensitivity to retrieval settings on LoCoMo with \texttt{Qwen3-8B}, where input tokens are per question relative to the default.}
\label{tab:ablation}
\vspace{0.5em}

\setlength{\tabcolsep}{3.5pt} 
\small

\begin{minipage}[t]{0.48\textwidth}
\centering
\begin{tabular}{lccc}
\toprule
Method & Judge & F1 & BLEU \\
\midrule
w/o context window & 62.87 & 43.56 & 38.49 \\
w/o pseudo-relevance feedback & 63.68 & 44.07 & 38.84 \\
w/o summary expansion & 64.46 & 44.29 & 39.12 \\
w/o wide first-stage scan & 63.94 & 44.53 & 39.28 \\
w/o summaries & 63.94 & 44.59 & 39.34 \\
w/o all (plain retrieval) & 59.09 & 42.10 & 36.99 \\
\midrule
\rowcolor{cyan!15}
\textbf{Full \name} & \textbf{64.59} & \textbf{45.42} & \textbf{39.92} \\
\bottomrule
\end{tabular}
\end{minipage}%
\hfill 
\begin{minipage}[t]{0.48\textwidth}
\centering
\begin{tabular}{lccc}
\toprule
Setting & Judge & F1 & Tokens \\
\midrule
Pack $k{=}5$ & 60.06 & 42.84 & $\times 0.56$ \\
\rowcolor{cyan!15}
Pack $k{=}10$ (default) & 61.82 & 44.19 & $2{,}156$ \\
Pack $k{=}20$ & 62.29 & 43.89 & $\times 1.85$ \\
Pack $k{=}30$ & 61.64 & 42.78 & $\times 2.65$ \\
\addlinespace[0.2em]
PRF, 4 documents & 61.23 & 43.99 & $\times 1.00$ \\
Summary expansion, 10 & 61.49 & 44.59 & $\times 1.01$ \\
Narrower pools ($30/40$) & $60.45$ & 43.26 & $\times 0.99$ \\
Wider scan ($100/50$) & 61.56 & 44.27 & $\times 1.01$ \\
\bottomrule
\end{tabular}
\end{minipage}
\end{table}

\subsection{Ablation study}
\label{sec:ablation}
We perform a systematic ablation study on the LoCoMo dataset to quantify the contribution of each core component of the \name{} architecture (Table~\ref{tab:ablation}, left). Removing the context window produces the largest degradation, with average F1 falling by 1.86 points and the judge score by 1.7, confirming that a retrieved turn is often not self-contained without its neighbors. Pseudo-relevance feedback, summary expansion, the wide first-stage scan, and the summaries themselves each account for between 0.83 and 1.35 average F1. The components therefore overlap, each recovering evidence the others partly reach, and removing all of them at once leaves plain retrieval, which costs 3.32 average F1 and 5.5 judge points ($p<0.001$). Appendix~\ref{sec:additional_results} reports per-category results.

\subsection{Sensitivity to retrieval settings}
\label{sec:sensitivity}
Table~\ref{tab:ablation} (right) changes one retrieval setting at a time, keeping the rest of the default configuration fixed. Across all of these settings, judge accuracy stays between 60.1\% and 62.3\%, so \name{} does not depend on careful tuning. The pack size shows the clearest trade-off: retrieving only $k{=}5$ candidates lowers average F1 by 1.35 points, while retrieving 20 or 30 brings no gain and makes each answer prompt up to $2.65\times$ longer. We therefore use $k{=}10$.

\section{Concluding remarks}
We present \name{}, a long-term agentic memory system that unifies efficient storage and effective retrieval. By drastically reducing LLM calls during memory construction while introducing more sophisticated retrieval, \name{} achieves strong performance while significantly reducing computational cost, specifically during memory insertion. Experiments across multiple benchmarks show that \name{} outperforms more expensive alternatives, while generalizing consistently across different model architectures and scales. These results indicate that \name{} provides a promising solution for real-time agentic systems.

\subsection*{AI use statement}

In this work, we used generative AI tools to assist with writing and debugging experiment code, organizing and checking experimental results, and editing the text of the paper for clarity. We did not use generative AI tools to generate research ideas or to produce experimental results. We have reviewed all AI-assisted work: code was checked and tested by the authors, every reported number was verified against the result files it comes from, and edited text was checked for accuracy against the experiments. We take responsibility for the final content of this work, including text, claims or artifacts produced with the aid of generative AI.

\subsection*{Ethics statement}

This work uses only publicly available benchmarks (LoCoMo, MemGallery, and LongMemEval-S), whose conversations are synthetic or were collected by their authors, and involves no new human subjects or personal data. Because \name{} stores conversations verbatim and never deletes on its own, a deployed system could retain sensitive information or stored injected content; we discuss these privacy and security risks, and the safeguards a deployment needs, in the broader impacts and limitations sections of the appendix. We are not aware of conflicts of interest or sponsorship related to this work.

\subsection*{Reproducibility statement}

We provide the code of \name{} and our LoCoMo evaluation on Github.~\footnote{\url{https://github.com/mitchellpiehl/MemFit}}
Section~\ref{sec:storage} and Section~\ref{sec:retrieval} describe the system, the implementation details in Section~4.1 list every setting, and Appendix~\ref{app:pseudocode} gives pseudocode for memory construction and retrieval. Appendix~\ref{sec:dataset} describes the datasets and evaluation protocols, Appendix~\ref{sec:prompt} lists every prompt, including the judge's, and Appendix~\ref{sec:additional_results} reports per-category results and the cost of every stage. All open-weight backbones are public models served by Ollama, and one configuration is used for every backbone and benchmark.

\subsection*{Acknowledgement}
This work was supported in part by the National Science Foundation (NSF) under Grant No. 2549600.

\bibliography{main}
\bibliographystyle{iclr2027_conference}

\appendix
\section{Pseudocode}
\label{app:pseudocode}

\begin{algorithm}[H]
\caption{Memory construction}
\label{alg:insertion}
\begin{algorithmic}[1]
\Require Utterance $(u_i, t_i, p_i, v_i)$; store $\mathcal{M}$ with BM25 index, embedding matrix $\mathbf{Z}$; encoder $f_{\text{enc}}$
\Ensure $\mathcal{M}$, the BM25 index, and $\mathbf{Z}$, each extended by one episode
\State $\mathbf{x}_i \gets u_i$ labeled with its time $t_i$ and speaker $p_i$
\If{$v_i$ is present} \Comment{Text-image fusion}
    \State Append the image identifier and caption of $v_i$ to $\mathbf{x}_i$ as structured text
\EndIf
\State $\mathbf{z}_i \gets f_{\text{enc}}(\mathbf{x}_i)$ \Comment{$L_2$-normalized}
\State $e_i \gets (\mathbf{x}_i, \mathbf{z}_i, g_i, \mathbf{m}_i)$; append $e_i$ to $\mathcal{M}$ \Comment{No LLM call}
\State Add the postings of $\mathbf{x}_i$ to the BM25 index; append $\mathbf{z}_i$ as a new row of $\mathbf{Z}$
\Statex
\Procedure{SummarizeSegments}{$\mathcal{M}$} \Comment{Off the query path}
    \State $b \gets \max(3, \mathrm{round}(\sqrt{N}/2))$
    \For{each gap $j$ between consecutive episodes}
        \State $c_j \gets \cos(\bar{\mathbf{z}}_{\text{left}}, \bar{\mathbf{z}}_{\text{right}})$ \Comment{Mean embeddings of the $b$ episodes on each side}
    \EndFor
    \State $\delta_j \gets$ depth of $c_j$ below its neighboring peaks, for every gap $j$
    \State Place a boundary at each local minimum of $c$ whose depth satisfies $\delta_j > \mu_\delta + \sigma_\delta/2$
    \For{each segment $S$ with $|S| \geq 3$} \Comment{At most $60$ per conversation}
        \State summary $\gets f(\text{up to } 14 \text{ evenly spaced excerpts of } S)$ \Comment{One LLM call; two sentences}
        \State Store the summary with its own embedding and with $S$ as its provenance
    \EndFor
\EndProcedure
\end{algorithmic}
\end{algorithm}

\begin{algorithm}[H]
\caption{Memory retrieval and answer extraction}
\label{alg:retrieval}
\begin{algorithmic}[1]
\Require Query $q$; store $\mathcal{M}$ with segment summaries; LLM $f$; encoder $f_{\text{enc}}$; cross-encoder $\mathrm{CE}$
\Ensure Answer $a$
\State $\mathbf{z}_q \gets f_{\text{enc}}(q)$
\State \Comment{\textbf{Stage 1: Hybrid candidate generation}}
\State $\mathcal{C} \gets$ top-$n_{\text{scan}}$ episodes by $\text{lex}(q, \mathbf{x}_i)$ $\cup$ top-$n_{\text{scan}}$ episodes by $\text{sem}(\mathbf{z}_q, \mathbf{z}_i)$
\State Score each $e_i \in \mathcal{C}$ by $s_i$ (Eq.~\ref{eq:fusion}), plus meta-data boosts; sort $\mathcal{C}$ by score
\State $\mathcal{S} \gets$ the $10$ summaries with the highest hybrid score
\State $\mathcal{R} \gets$ the first $50$ entries of $\mathcal{C}$ and $\mathcal{S}$, interleaved by rank \Comment{Summaries compete with episodes}
\State \Comment{\textbf{Stage 2: Additive candidate expansion}}
\State $\mathcal{E} \gets$ member episodes of the $5$ highest-scoring summaries in $\mathcal{S}$ \Comment{Provenance}
\State $\mathcal{E} \gets \mathcal{E} \cup$ Stage 1 rerun with the $12$ rarest terms of the top-$2$ episodes in $\mathcal{R}$ \Comment{Pseudo-relevance feedback}
\State \Comment{\textbf{Stage 3: Reranking and pack assembly}}
\State $\mathcal{P} \gets \mathcal{R} \cup \mathcal{E}$ \Comment{Expansion adds, never evicts}
\State Score every candidate in $\mathcal{P}$ against $q$ with $\mathrm{CE}$; standardize the scores within episodes and within summaries
\State $\mathcal{H}_k \gets$ top-$k$ of $\mathcal{P}$, each episode with its neighbors within $w$ in the same session
\State \Comment{\textbf{Stage 4: Answer extraction}}
\State $a \gets f(q, \mathcal{H}_k)$ \Comment{The only LLM call: chronological excerpts, dates resolved}
\State \Return $a$
\end{algorithmic}
\end{algorithm}

\section{Baselines and datasets}

\subsection{Datasets}
\label{sec:dataset}
Table~\ref{tab:datasets} summarizes the three benchmarks. All three are stored the same way: one episode per turn, labeled with its date and speaker, followed by segment summaries, with nothing tuned per benchmark.

\begin{table}[h]
\centering
\caption{Benchmark statistics.}
\label{tab:datasets}
\small
\begin{tabular}{@{}lccc@{}}
\toprule
 & LoCoMo & MemGallery & LongMemEval-S \\
\midrule
Questions evaluated & 1,540 of 1,986 & 1,711 & 500 \\
Question categories & 4 (+ adversarial) & 9 & 6 (+ abstention) \\
Metrics & F1, BLEU-1, judge & F1, judge & judge (official prompts) \\
\bottomrule
\end{tabular}
\end{table}

\paragraph{LoCoMo.}
LoCoMo \citep{maharana2024evaluating} contains 10 conversations between two speakers, each spanning 19 to 32 sessions over several months, with 5,882 turns in total. Of its 1,986 questions, we evaluate the 1,540 non-adversarial ones, following prior work:
\begin{itemize}[leftmargin=*]
    \item \textbf{Multi-Hop (282):} Requires the model to synthesize and compose information across multiple conversational sessions.
    \item \textbf{Temporal (321):} Involves date arithmetic, event ordering, and reasoning about the timing of specific occurrences.
    \item \textbf{Open Domain (96):} Requires combining interaction-specific details with world knowledge and preferences established during the conversation.
    \item \textbf{Single-Hop (841):} Focuses on factual recall that is local to a single session.
    \item \textbf{Adversarial (446, excluded):} Includes unanswerable distractors and trick questions that test refusal.
\end{itemize}

\paragraph{MemGallery.}
MemGallery \citep{bei2026mem} contains 20 scenarios, each a multi-session conversation between one user persona and an assistant, with 240 sessions and 3,962 rounds in total; 1,003 rounds share an image, and the dataset provides a caption for each. Each round becomes one episode holding the user message, the assistant reply, and the identifier and caption of its image. The benchmark's 1,711 questions, 487 of which come with an image of their own, fall into nine categories grouped into three dimensions:
\begin{itemize}[leftmargin=*]
    \item \textbf{Memory extraction and adaptation.} \emph{Factual Retrieval} (FR, 219): recall explicit textual information from earlier in the dialogue. \emph{Visual-centric Search} (VS, 306): find specific images in a long multimodal history. \emph{Test-Time Learning} (TTL, 337): apply visual or textual knowledge introduced during the conversation; 333 of these questions ask about an image they come with.
    \item \textbf{Memory reasoning.} \emph{Temporal Reasoning} (TR, 123): resolve time-dependent questions using multimodal clues. \emph{Visual-centric Reasoning} (VR, 174): infer an answer from visual cues in the conversation's images. \emph{Multi-entity Reasoning} (MR, 206): combine several retrieved entities or facts.
    \item \textbf{Memory knowledge management.} \emph{Knowledge Resolution} (KR, 81): prefer the most recent of several conflicting statements. \emph{Conflict Detection} (CD, 81): identify a contradiction with stored memory. \emph{Answer Refusal} (AR, 184): decline when the information was never given.
\end{itemize}
We follow the benchmark's reference implementation: its system prompt is prepended to the reader prompt, the official answer-format instruction of the AR, CD, and VS categories is appended to their questions, and the reader's refusal is reported as the benchmark's refusal string, ``Not mentioned.'' A question's own image is described by its caption in the query; in the image-bearing configurations, the image itself is also passed to the reader. We report token-level F1 over all 1,711 questions (the benchmark's Total) and LLM-judge accuracy.

\paragraph{LongMemEval-S.}
LongMemEval-S \citep{wu2024longmemeval} contains 500 questions, each with its own timestamped history of chat sessions between a user and an assistant: 47.7 sessions and roughly 115k tokens per history, with the sessions that hold the evidence embedded among filler sessions. We use the cleaned release published by the benchmark's authors. The questions fall into six types, and 30 of them are abstention questions (6 SSU, 12 MS, 6 KU, 6 TR) about things that were never mentioned, which the system should decline to answer:
\begin{itemize}[leftmargin=*]
    \item \textbf{Single-session-user} (SSU, 70): recall a detail the user stated in one session.
    \item \textbf{Single-session-assistant} (SSA, 56): recall something the assistant said in one session.
    \item \textbf{Single-session-preference} (SSP, 30): give a response that reflects a preference the user expressed, graded against a rubric.
    \item \textbf{Multi-session} (MS, 133): aggregate or compare information spread across several sessions.
    \item \textbf{Knowledge-update} (KU, 78): report the latest value of user information that changed over time.
    \item \textbf{Temporal-reasoning} (TR, 133): reason about dates, durations, and order from session timestamps and time mentions.
\end{itemize}
Each history is stored exactly as a LoCoMo conversation: one episode per message, from the user and the assistant alike, since SSA questions ask what the assistant said, in chronological order with each session's timestamp, followed by segment summaries written by the backbone. As in the benchmark's own reader prompt, the question is followed by the date on which it is asked (``Current date: \dots''). Because this is the only benchmark that dates its questions, the reader here also sees each excerpt's date as its distance from that date (``10 days before the current date''), and relative time expressions in the question are resolved against it (``two weeks ago (around 16 May 2023)''). Both are rule-based, like the resolution of dates in excerpts, and retrieval sees the question as asked. A NOT\_FOUND reply is reported as ``The information is not available in our conversation history.'' Answers are graded with the official evaluation prompts, one per question type plus one for abstention. The comparison in Table~\ref{tab:overall} uses \texttt{GPT-4o-mini} (\texttt{gpt-4o-mini-2024-07-18}) as both backbone and judge, the protocol of the LightMem results we compare against \citep{fang2025lightmem}; a second run with \texttt{Qwen3-8B} uses the official judge, \texttt{GPT-4o} (\texttt{gpt-4o-2024-08-06}).

\subsection{Comparative baselines}
\label{sec:baseline}
Table~\ref{tab:baselines} lists the systems we compare against and where each appears. On MemGallery, baseline results are those published by the benchmark's authors \citep{bei2026mem} with \texttt{Qwen2.5-VL-7B} under the official protocol. On LongMemEval-S, they are those reported by \citet{fang2025lightmem} with \texttt{GPT-4o-mini} as backbone and judge. The efficiency measurements in Figure~\ref{fig:efficiency_and_latency} are our own, on the same benchmark and backbone as \name{}; every LLM call their released code sends is metered at the HTTP layer.

\begin{table}[h]
\centering
\caption{Baselines, what spends LLM calls when each builds memory, and where each is compared: LoCoMo (Table~\ref{tab:locomo1}), MemGallery and LongMemEval-S (Table~\ref{tab:overall}), and efficiency (Figure~\ref{fig:efficiency_and_latency}).}
\label{tab:baselines}
\small
\begin{tabular}{@{}llcccc@{}}
\toprule
System & LLM work at write time & LoCoMo & MemGallery & LongMemEval-S & Efficiency \\
\midrule
Full context & none (no memory) & & \checkmark & \checkmark & \\
NaiveRAG & none (embedded chunks) & & \checkmark & \checkmark & \\
MemoryBank & summaries, memory decay & \checkmark & & & \\
MemGPT & LLM-managed memory tiers & \checkmark & \checkmark & & \\
MemoryOS & LLM-updated memory tiers & & \checkmark & \checkmark & \\
A-MEM & notes, links, note evolution & \checkmark & \checkmark & \checkmark & \checkmark \\
Mem0 & fact extraction and update & \checkmark & & \checkmark & \checkmark \\
LangMem & memory extraction and update & & & \checkmark & \\
LightMem & topic summaries and update & \checkmark & & \checkmark & \checkmark \\
SimpleMem & compression to memory units & \checkmark & & & \checkmark \\
\midrule
\rowcolor{cyan!20}
\textbf{\name} & one summary per segment & \checkmark & \checkmark & \checkmark & \checkmark \\
\bottomrule
\end{tabular}
\end{table}

\paragraph{Full context.} The entire history is placed in the reader's context window, with no memory system. This measures how well the backbone reads a long history on its own.

\paragraph{NaiveRAG.} A standard retrieval-augmented baseline: the history is split into chunks, the chunks are embedded, and those most similar to the question are passed to the reader. Nothing is summarized, linked, or rewritten.

\paragraph{MemoryBank} \citep{zhong2024memorybank}.
MemoryBank implements a memory management framework based on the Ebbinghaus forgetting curve. It stores past interactions in a vector database and reinforces or decays memories according to their temporal significance and how often they are accessed, so that the agent keeps a consistent persona and recalls behavioral associations relevant to the current dialogue.

\paragraph{MemGPT} \citep{packer2023memgpt}.
Inspired by operating systems, MemGPT manages a ``virtual context'' by treating the LLM's fixed context window as a tiered memory hierarchy, with a main context (analogous to RAM) and an external context (analogous to disk). The LLM manages its own memory through explicit function calls that move information between the tiers.

\paragraph{MemoryOS} \citep{kang2025memory}.
MemoryOS applies operating-system memory management to conversational agents. It stores memory in three tiers, short-term, mid-term, and long-term personal memory, and moves content between them with LLM-driven updates: dialogue pages pass from short-term to mid-term memory first-in-first-out and are grouped into topic segments, and frequently accessed segments are promoted to long-term memory, which maintains user and agent profiles.

\paragraph{A-MEM} \citep{xu2025mem}.
A-MEM is an agentic memory framework inspired by the Zettelkasten method. It represents knowledge as a network of atomic notes connected by semantic links. When a memory is added, an LLM writes its note, links it to related notes, and may update those notes (``memory evolution''), so that the agent maintains an interconnected web of knowledge rather than a list of interaction logs.

\paragraph{Mem0} \citep{chhikara2025mem0}.
Mem0 maintains a store of salient facts. As messages arrive, an LLM extracts candidate facts, compares each with similar stored memories, and chooses to add, update, or delete memories, or leave the store unchanged; a graph-based variant additionally stores relations between entities.

\paragraph{LangMem.} LangMem is LangChain's long-term memory library. An LLM extracts memories from conversations and consolidates them by creating, updating, or deleting stored entries, which are retrieved by semantic search.

\paragraph{LightMem} \citep{fang2025lightmem}.
Following the Atkinson--Shiffrin model of human memory, LightMem partitions memory into sensory, short-term, and long-term stages. A pre-compression step filters redundant tokens before information enters memory, topic-aware segments are summarized into memory entries, and memory organization and updates run offline in a ``sleep-time'' process that is decoupled from inference.

\paragraph{SimpleMem} \citep{liu2026simplemem}.
SimpleMem focuses on semantic lossless compression. It uses an entropy-aware filtering mechanism to distill unstructured interactions into compact, context-independent memory units indexed from multiple views, reducing the token footprint of retrieving from long interaction histories.

\section{Additional results}
\label{sec:additional_results}
This section reports per-category results for every configuration in the paper, together with results that did not fit in the main text. Unless stated otherwise, every run uses the configuration described in the main text, with one LLM call per question. LLM-judge accuracy on LoCoMo and MemGallery is scored by \texttt{gpt-oss-20b} with the prompt in Appendix~\ref{sec:prompt}. Differences between two configurations are tested with a paired sign-flip permutation test over the per-question judge verdicts, and ``gained/lost'' counts the questions that one configuration answers correctly and the other does not.

\subsection{LoCoMo}

\paragraph{Plain retrieval against \name{} on five readers.}
Table~\ref{tab:lc_readers} compares \name{} with plain retrieval over the same store: the same encoders, cross-encoder, and reader, without summaries, expansions, the context window, or the wide scan. In addition to the four backbones of Table~\ref{tab:locomo1}, it includes \texttt{Gemma4-31B}. The memory design adds 4.6 to 7.9 judge points on every reader ($p<0.001$) and 2.8 to 4.5 average F1.

\begin{table}[h]
\centering
\caption{Plain retrieval against \name{} on LoCoMo with five readers. F1 and BLEU are averaged over the four categories, as in Table~\ref{tab:locomo1}; gained/lost counts the questions \name{} answers correctly and plain retrieval does not, and vice versa.}
\label{tab:lc_readers}
\small
\begin{tabular}{@{}lcccccc@{}}
\toprule
 & \multicolumn{2}{c}{Plain retrieval} & \multicolumn{2}{c}{\name} & \multicolumn{2}{c}{Judge difference} \\
\cmidrule(lr){2-3} \cmidrule(lr){4-5} \cmidrule(l){6-7}
Reader & Judge & F1 / BLEU & Judge & F1 / BLEU & $\Delta$ ($p$) & Gained/lost \\
\midrule
\texttt{Qwen3-8B} & 57.24 & 41.41 / 36.81 & 61.88 & 44.20 / 39.38 & $+4.61$ ($<$0.001) & 178/107 \\
\texttt{Gemma3-27B} & 59.09 & 42.10 / 36.99 & 64.59 & 45.42 / 39.92 & $+5.52$ ($<$0.001) & 160/75 \\
\texttt{Gemma4-31B} & 61.75 & 45.11 / 40.48 & 69.42 & 48.55 / 43.47 & $+7.66$ ($<$0.001) & 160/42 \\
\texttt{GPT-4.1-mini} & 64.33 & 47.30 / 42.10 & 72.37 & 50.26 / 44.74 & $+7.94$ ($<$0.001) & 185/63 \\
\texttt{GPT-4o} & 60.06 & 44.87 / 39.58 & 64.91 & 49.33 / 43.96 & $+4.81$ ($<$0.001) & 153/79 \\
\bottomrule
\end{tabular}
\end{table}

\paragraph{Ablation per category and across readers.}
Table~\ref{tab:lc_ablation_cat} expands the ablation of Table~\ref{tab:ablation} to every category. Every component helps open-domain questions (18.42 F1 with all components, 15.14 to 17.09 without any one of them), and multi-hop questions lose 1.0 to 1.6 F1 without any single component and 3.8 with plain retrieval. The context window carries most of the single-hop gain (63.88 to 59.63 F1), since the answer to a retrieved question often sits in the reply that follows it. Temporal questions barely move (58.05 to 59.66 F1). Open-domain has only 96 questions, so its per-component differences are noisy. Table~\ref{tab:lc_ablation_readers} repeats the ablation on \texttt{Qwen3-8B} and \texttt{Gemma4-31B}. Removing any component lowers judge accuracy on all three readers, in 18 of 18 comparisons, although individually only the context window (on two readers) and the wide scan (on one) reach $p<0.05$, besides plain retrieval on all three. Retrieval shows why: each component raises the share of questions with gold evidence among the top 30 candidates, from 86.7--88.0\% without it to 88.5\% with every component, against 83.4\% for plain retrieval.

\begin{table}[h]
\centering
\caption{Ablation of \name{} per category on LoCoMo with \texttt{Gemma3-27B}: F1 and BLEU-1 (\%) per category, their four-category average, and LLM-judge accuracy, with the judge difference from the full system and its $p$-value.}
\label{tab:lc_ablation_cat}
\resizebox{\textwidth}{!}{
\renewcommand{\arraystretch}{1.15}
\begin{tabular}{l|cccccccc|cc|cc}
\toprule
\multirow{2}{*}{Method} & \multicolumn{2}{c}{Multi-Hop} & \multicolumn{2}{c}{Temporal} & \multicolumn{2}{c}{OpenDomain} & \multicolumn{2}{c|}{Single-Hop} & \multicolumn{2}{c|}{Average} & \multicolumn{2}{c}{Judge} \\
\cmidrule(lr){2-3} \cmidrule(lr){4-5} \cmidrule(lr){6-7} \cmidrule(lr){8-9} \cmidrule(lr){10-11} \cmidrule(lr){12-13}
& F1 & BLEU & F1 & BLEU & F1 & BLEU & F1 & BLEU & F1 & BLEU & Acc. & $\Delta$ ($p$) \\
\midrule
w/o context window & 38.85 & 31.81 & 58.67 & 53.26 & 17.09 & 14.96 & 59.63 & 53.92 & 43.56 & 38.49 & 62.87 & $-1.69$ (0.054) \\
w/o pseudo-relevance feedback & 39.40 & 32.47 & 58.05 & 52.21 & 15.72 & 13.11 & 63.09 & 57.55 & 44.07 & 38.84 & 63.68 & $-0.98$ (0.138) \\
w/o summary expansion & 39.36 & 32.22 & 59.18 & 53.36 & 15.14 & 13.08 & 63.50 & 57.83 & 44.29 & 39.12 & 64.46 & $-0.13$ (0.884) \\
w/o wide first-stage scan & 39.28 & 31.85 & 58.90 & 53.20 & 16.15 & 14.07 & 63.78 & 58.01 & 44.53 & 39.28 & 63.94 & $-0.72$ (0.271) \\
w/o summaries & 38.95 & 32.10 & \textbf{59.66} & \textbf{53.84} & 15.66 & 13.07 & \textbf{64.09} & \textbf{58.37} & 44.59 & 39.34 & 63.94 & $-0.65$ (0.418) \\
\midrule
\rowcolor{cyan!20}
\textbf{Full \name} & \textbf{40.45} & \textbf{33.27} & 58.93 & 53.18 & \textbf{18.42} & \textbf{15.14} & 63.88 & 58.11 & \textbf{45.42} & \textbf{39.92} & \textbf{64.59} & -- \\
\bottomrule
\end{tabular}
}
\end{table}

\begin{table}[h]
\centering
\caption{Ablation on LoCoMo across three readers: judge difference from the full system ($p$-value), and retrieval measured independently of the reader. Pack hit is the share of questions with at least one gold evidence turn among the $k$ retrieved items; hit@30 is the same among the top 30 reranked candidates. Removing the context window leaves retrieval unchanged because neighbors are added after ranking.}
\label{tab:lc_ablation_readers}
\small
\begin{tabular}{@{}lccccc@{}}
\toprule
 & \multicolumn{3}{c}{Judge $\Delta$ ($p$)} & \multicolumn{2}{c}{Retrieval (\%)} \\
\cmidrule(lr){2-4} \cmidrule(l){5-6}
Configuration & \texttt{Qwen3-8B} & \texttt{Gemma3-27B} & \texttt{Gemma4-31B} & Pack hit & Hit@30 \\
\midrule
w/o context window & $-2.53$ (0.008) & $-1.69$ (0.054) & $-3.51$ ($<$0.001) & 82.5 & 88.5 \\
w/o pseudo-relevance feedback & $-1.23$ (0.082) & $-0.98$ (0.138) & $-0.58$ (0.340) & 81.5 & 88.0 \\
w/o summary expansion & $-0.91$ (0.111) & $-0.13$ (0.884) & $-0.20$ (0.744) & 80.9 & 86.7 \\
w/o wide first-stage scan & $-1.43$ (0.034) & $-0.72$ (0.271) & $-0.58$ (0.325) & 81.1 & 87.5 \\
w/o summaries & $-1.56$ (0.055) & $-0.65$ (0.418) & $-0.71$ (0.222) & 82.1 & 87.4 \\
w/o all (plain retrieval) & $-4.61$ ($<$0.001) & $-5.52$ ($<$0.001) & $-7.66$ ($<$0.001) & 79.2 & 83.4 \\
\midrule
\rowcolor{cyan!20}
\textbf{Full \name{}} (judge accuracy) & 61.88 & 64.59 & 69.42 & 82.5 & 88.5 \\
\bottomrule
\end{tabular}
\end{table}

\paragraph{Where the retrieval gains come from.}
Because retrieval does not involve the reader, Table~\ref{tab:lc_retrieval} holds for every backbone. \name{} raises the share of questions whose pack contains gold evidence from 79.2\% to 82.5\%, and the share with gold evidence among the top 30 candidates from 83.4\% to 88.5\%, with the largest gains on multi-hop (74.8\% to 81.2\% pack hit) and open-domain questions (51.1\% to 57.6\%). Multi-hop questions remain the hardest: the pack holds every one of their evidence turns for only 26.2\% of them (23.0\% with plain retrieval), which is where systems that link memories at write time can gain on \name{}.

\begin{table}[h]
\centering
\caption{Evidence retrieval on LoCoMo per category, plain retrieval (Plain) against \name{}, over the 1,536 questions with annotated evidence. Pack hit: at least one gold turn among the $k$ retrieved items. Pack all: every gold turn among them. Hit@30: at least one gold turn among the top 30 reranked candidates.}
\label{tab:lc_retrieval}
\small
\begin{tabular}{@{}lcccccc@{}}
\toprule
 & \multicolumn{2}{c}{Pack hit (\%)} & \multicolumn{2}{c}{Pack all (\%)} & \multicolumn{2}{c}{Hit@30 (\%)} \\
\cmidrule(lr){2-3} \cmidrule(lr){4-5} \cmidrule(l){6-7}
Category & Plain & \name & Plain & \name & Plain & \name \\
\midrule
Multi-Hop & 74.8 & \textbf{81.2} & 23.0 & \textbf{26.2} & 81.6 & \textbf{87.6} \\
Temporal & 83.5 & \textbf{85.7} & 77.9 & \textbf{79.8} & 85.0 & \textbf{89.4} \\
OpenDomain & 51.1 & \textbf{57.6} & 34.8 & \textbf{39.1} & 62.0 & \textbf{68.5} \\
Single-Hop & 82.2 & \textbf{84.4} & 79.5 & \textbf{81.9} & 85.7 & \textbf{90.7} \\
\midrule
All & 79.2 & \textbf{82.5} & 66.1 & \textbf{68.7} & 83.4 & \textbf{88.5} \\
\bottomrule
\end{tabular}
\end{table}

\paragraph{Sensitivity per category.}
Table~\ref{tab:lc_sensitivity} gives every setting of the sensitivity study. On \texttt{Gemma3-27B}, the pack size increasing leads to better performance. Specifically, $k{=}5$ costs 3.12 judge points ($p<0.001$), while $k{=}20$ and $k{=}30$ change the judge score by $+0.85$ ($p{=}0.37$) and $-0.13$ ($p{=}0.95$) at $1.85\times$ and $2.65\times$ the input tokens.

\begin{table}[h]
\centering
\caption{Sensitivity of \name{} on LoCoMo, one setting at a time: per-category F1, four-category average F1 and BLEU-1, LLM-judge accuracy with its difference from the default, and input tokens per question relative to the default. Pools are given as $n_{\text{scan}}$/number of first-stage candidates reranked.}
\label{tab:lc_sensitivity}
\resizebox{\textwidth}{!}{
\renewcommand{\arraystretch}{1.12}
\begin{tabular}{l|cccc|cc|cc|c}
\toprule
Setting & Multi-Hop & Temporal & OpenDomain & Single-Hop & Avg.\ F1 & Avg.\ BLEU & Judge & $\Delta$ ($p$) & Tokens \\
\midrule
\multicolumn{10}{l}{\emph{\texttt{Qwen3-8B}}} \\
\rowcolor{cyan!20}
Default ($k{=}10$, $w{=}1$, PRF 2, expansion 5, pools $60/50$) & 40.22 & 59.26 & 16.66 & 60.63 & 44.19 & 39.38 & 61.82 & -- & 2{,}156 \\
Pack $k{=}5$ & 34.73 & 57.75 & 17.83 & 61.07 & 42.84 & 38.12 & 60.06 & $-1.75$ (0.081) & $\times 0.56$ \\
Pack $k{=}20$ & 38.14 & 59.75 & 16.61 & 61.08 & 43.89 & 39.12 & 62.29 & $+0.46$ (0.666) & $\times 1.85$ \\
Pack $k{=}30$ & 37.25 & 56.84 & 16.82 & 60.21 & 42.78 & 38.15 & 61.64 & $-0.13$ (0.948) & $\times 2.65$ \\
Context window $w{=}2$ & 38.52 & 60.38 & 20.95 & 62.80 & 45.66 & 40.76 & 63.68 & $+1.82$ (0.036) & $\times 1.44$ \\
PRF, 4 documents & 39.31 & 60.24 & 15.94 & 60.46 & 43.99 & 39.08 & 61.23 & $-0.58$ (0.225) & $\times 1.00$ \\
Summary expansion, 10 & 39.83 & 61.38 & 16.64 & 60.52 & 44.59 & 39.77 & 61.49 & $-0.32$ (0.572) & $\times 1.01$ \\
Narrower pools ($30/40$) & 36.98 & 58.57 & 17.26 & 60.25 & 43.26 & 38.59 & 60.45 & $-1.36$ (0.044) & $\times 0.99$ \\
Wider scan ($100/50$) & 39.13 & 60.96 & 15.99 & 61.02 & 44.27 & 39.56 & 61.56 & $-0.26$ (0.719) & $\times 1.01$ \\
\midrule
\multicolumn{10}{l}{\emph{\texttt{Gemma3-27B}}} \\
\rowcolor{cyan!20}
Default ($k{=}10$) & 40.45 & 58.93 & 18.42 & 63.88 & 45.42 & 39.92 & 64.59 & -- & 2{,}181 \\
Pack $k{=}5$ & 34.92 & 57.56 & 17.05 & 62.42 & 42.99 & 37.67 & 61.49 & $-3.12$ ($<$0.001) & $\times 0.56$ \\
Pack $k{=}20$ & 41.68 & 60.79 & 15.06 & 65.00 & 45.63 & 40.32 & 65.54 & $+0.85$ (0.373) & $\times 1.85$ \\
Pack $k{=}30$ & 41.02 & 59.94 & 16.79 & 65.23 & 45.75 & 40.40 & 64.48 & $-0.13$ (0.945) & $\times 2.65$ \\
\bottomrule
\end{tabular}
}
\end{table}

\subsection{MemGallery and LongMemEval-S}

Table~\ref{tab:memgallery_f1} gives the per-category F1 behind the MemGallery column of Table~\ref{tab:overall}. Passing the question's own image to the reader adds 2.85 F1 over text alone (64.82 to 67.66; judge $+2.63$, $p<0.001$), concentrated in Test-Time Learning (72.7 to 82.4), where 333 of 337 questions ask about the image they come with, and Visual-centric Reasoning (45.2 to 52.7). Also passing the images of the retrieved episodes adds nothing further overall (67.47), trading Test-Time Learning for Knowledge Resolution and Visual-centric Search. As a text-only reader, \texttt{Gemma3-27B} reaches 69.85 F1.

\begin{table}[h]
\centering
\caption{Token-level F1 (\%) on MemGallery per category under the benchmark's official protocol. Baseline rows are as published \citep{bei2026mem}; \name{} per-category values are given to one decimal. The upper block uses \texttt{Qwen2.5-VL-7B} throughout; the lower block runs \name{} as a text-only system with other readers. Best result in each column of the upper block in \textbf{bold}.}
\label{tab:memgallery_f1}
\resizebox{\textwidth}{!}{
\renewcommand{\arraystretch}{1.15}
\begin{tabular}{l|ccccccccc|c}
\toprule
Method & AR & CD & FR & KR & MR & TR & TTL & VR & VS & Total \\
\midrule
NaiveRAG & 95.81 & 34.57 & 58.52 & 35.95 & 46.40 & 48.87 & 65.26 & 30.22 & 75.92 & 59.74 \\
MemGPT & \textbf{98.49} & \textbf{35.80} & 59.28 & 42.92 & 43.67 & 56.61 & 29.24 & 45.93 & 62.39 & 52.82 \\
A-MEM & 92.78 & 33.33 & 60.72 & 45.15 & \textbf{50.00} & 56.04 & 63.36 & 44.77 & 76.81 & 62.28 \\
MemoryOS & 98.45 & 33.33 & 62.44 & \textbf{50.31} & 44.90 & 54.97 & 54.84 & 42.80 & 78.53 & 61.09 \\
\midrule
\name{}, plain retrieval, text only & 93.5 & \textbf{35.8} & 65.7 & 38.9 & 40.9 & 65.6 & 71.8 & 41.4 & 77.2 & 63.79 \\
\name{}, text only & 94.6 & \textbf{35.8} & 67.6 & 38.9 & 42.9 & 69.5 & 72.7 & 45.2 & 74.8 & 64.82 \\
\rowcolor{cyan!20}
\textbf{\name} + question image & 94.6 & 33.3 & \textbf{68.0} & 39.2 & 42.6 & 69.5 & \textbf{82.4} & 52.7 & 76.3 & \textbf{67.66} \\
\name{} + question and retrieved images & 94.6 & \textbf{35.8} & 66.6 & 45.9 & 41.5 & \textbf{69.7} & 76.9 & \textbf{54.7} & \textbf{79.4} & 67.47 \\
\midrule
\name{}, text only, \texttt{Qwen3-8B} & 78.4 & 45.7 & 69.9 & 44.7 & 49.0 & 57.7 & 73.3 & 57.6 & 57.1 & 62.22 \\
\name{}, text only, \texttt{Gemma3-27B} & 65.3 & 65.4 & 69.3 & 51.4 & 47.5 & 72.9 & 78.5 & 63.3 & 87.0 & 69.85 \\
\bottomrule
\end{tabular}
}
\end{table}

Table~\ref{tab:memgallery_more} shows that the ablation is flat on MemGallery. No component changes judge accuracy significantly, and with \texttt{Qwen2.5-VL-7B} plain retrieval is within 1.52 judge points of the full system ($p{=}0.094$). Retrieval explains why: MemGallery's scenarios hold about 200 rounds each, and plain retrieval already places gold evidence in the pack for 95.5\% of questions, so the expansions have little left to recover, and accuracy is limited by the reader. With \texttt{Qwen3-8B}, the gain over plain retrieval is significant ($+3.27$ judge points, $p{=}0.001$). With one category-blind format instruction given to every question instead of the per-category instructions of the reference implementation, F1 falls from 64.82 to 47.92. All published results in Table~\ref{tab:memgallery_f1} use the official protocol. For LongMemEval-S, Table \ref{tab:lme_cost} shows MemFit issues 40 LLM calls per LongMemEval-S history, 13–73× fewer than LangMem, Mem0, A-MEM, and MemoryOS, and fewer tokens than LightMem.

\begin{table}[h]
\centering
\caption{MemGallery under the official protocol: readers, images, ablation, and protocol. F1 and LLM-judge accuracy (\%), with the judge difference from the text-only full system on the same reader. Text only unless images are listed.}
\label{tab:memgallery_more}
\small
\begin{tabular}{@{}llccc@{}}
\toprule
Reader & Configuration & F1 & Judge & $\Delta$ judge ($p$) \\
\midrule
\multirow{11}{*}{\texttt{Qwen2.5-VL-7B}} & \name{} & 64.82 & 64.23 & -- \\
 & \quad + question image & \textbf{67.66} & \textbf{66.86} & $+2.63$ ($<$0.001) \\
 & \quad + question and retrieved images & 67.47 & 66.63 & $+2.40$ (0.010) \\
 & \quad w/o context window & 64.91 & 64.00 & $-0.23$ (0.827) \\
 & \quad w/o pseudo-relevance feedback & 65.12 & 64.76 & $+0.53$ (0.218) \\
 & \quad w/o summary expansion & 65.07 & 64.35 & $+0.12$ (0.805) \\
 & \quad w/o wide first-stage scan & 65.13 & 64.87 & $+0.64$ (0.114) \\
 & \quad w/o summaries & 63.96 & 63.00 & $-1.23$ (0.069) \\
 & \quad context window $w{=}2$ & 65.15 & 64.29 & $+0.06$ (1.000) \\
 & \quad plain retrieval & 63.79 & 62.71 & $-1.52$ (0.094) \\
 & \quad category-blind format instruction & 47.92 & 47.75 & $-16.48$ ($<$0.001) \\
\midrule
\multirow{2}{*}{\texttt{Qwen3-8B}} & \name{} & 62.22 & 68.95 & -- \\
 & \quad plain retrieval & 59.44 & 65.69 & $-3.27$ (0.001) \\
\midrule
\texttt{Gemma3-27B} & \name{}, own summaries & 69.85 & 71.30 & -- \\
\bottomrule
\end{tabular}
\end{table}

\begin{table}[h]
\centering
\caption{Memory construction cost per LongMemEval-S history, excluding the answer call, with accuracy (\texttt{GPT-4o-mini} backbone and judge). Baseline values are from Table~2 of \citet{fang2025lightmem}, measured with \texttt{GPT-4o-mini}. \name{}'s calls do not depend on the backbone, since segmentation uses no LLM; its tokens were measured with \texttt{Qwen3-8B}.}
\label{tab:lme_cost}
\small
\begin{tabular}{@{}lccc@{}}
\toprule
Method & LLM calls & Tokens (k) & Accuracy (\%) \\
\midrule
LangMem & 520.62 & 1,102.16 & 37.20 \\
A-MEM & 986.55 & 1,605.81 & 62.60 \\
MemoryOS & 2,938.41 & 2,991.75 & 44.80 \\
Mem0 & 811.57 & 1,152.62 & 53.61 \\
LightMem & \textbf{25.67} & 30.81 & 64.29 \\
\rowcolor{cyan!20}
\textbf{\name} & 40.14 & \textbf{25.07} & \textbf{64.60} \\
\bottomrule
\end{tabular}
\end{table}

\subsection{Cost of every stage}

Table~\ref{tab:stage_cost} breaks the cost of a complete run into stages, for two backbones on LoCoMo and MemGallery, with each model writing its own summaries. Storing and indexing the turns takes 15\,s for all of LoCoMo and 18\,s for all of MemGallery, and segmenting them takes under 0.2\,s; the rest of the build is the summary calls. At query time, retrieval takes 87 to 112\,ms per question, about 60--65\% of it in the cross-encoder, and the answer call takes the rest. Across readers, \name{}'s answer prompt is $2.07$--$2.10\times$ as long as plain retrieval's on LoCoMo and $1.64\times$ on MemGallery, the price of the context window and the summaries.

\begin{table}[h]
\centering
\caption{Cost of every stage of a complete run: memory construction for every conversation or scenario, run serially on one GPU, and answering every question, run with eight model instances on four GPUs. Latency is per question, including the answer call.}
\label{tab:stage_cost}
\resizebox{\textwidth}{!}{
\renewcommand{\arraystretch}{1.12}
\begin{tabular}{l|cc|cc}
\toprule
 & \multicolumn{2}{c|}{LoCoMo (10 conversations, 1,540 questions)} & \multicolumn{2}{c}{MemGallery (20 scenarios, 1,711 questions)} \\
Stage & \texttt{Qwen3-8B} & \texttt{Gemma3-27B} & \texttt{Qwen3-8B} & \texttt{Gemma3-27B} \\
\midrule
\multicolumn{5}{l}{\emph{Build (once, before any question)}} \\
Embed turns (s) & 1.18 & 1.19 & 1.51 & 1.49 \\
Index and insert (s) & 13.83 & 13.81 & 16.58 & 16.57 \\
Segment with TextTiling (s) & 0.18 & 0.16 & 0.16 & 0.16 \\
Summary LLM calls & 437 & 437 & 287 & 287 \\
Summary tokens, in / out & 172,180 / 27,248 & 177,460 / 44,878 & 173,431 / 20,322 & 174,788 / 34,453 \\
Summarize (s) & 236.3 & 924.1 & 174.0 & 711.6 \\
Embed summaries (s) & 4.20 & 6.80 & 2.91 & 4.30 \\
Build total (s) & 255.7 & 946.2 & 195.2 & 734.2 \\
\midrule
\multicolumn{5}{l}{\emph{Query (per question)}} \\
LLM calls & 1.00 & 1.00 & 1.00 & 1.00 \\
Answer tokens, in / out & 2,156 / 9.7 & 2,210 / 8.7 & 3,208 / 13.4 & 3,298 / 10.4 \\
Cross-encoder pairs & 188 & 186 & 137 & 137 \\
Retrieval latency (s) & 0.087 & 0.105 & 0.092 & 0.112 \\
Latency, mean / p95 (s) & 0.67 / 0.85 & 1.95 / 2.46 & 0.81 / 1.21 & 2.58 / 3.59 \\
\midrule
Judge accuracy (\%) & 61.88 & 64.37 & 68.83 & 71.30 \\
\bottomrule
\end{tabular}
}
\end{table}
\section{Broader impacts}
\paragraph{Positive societal contributions.} By enabling efficient, long-term memory, \name{} advances the development of agents capable of maintaining context over extended interactions. This technology has the potential to improve accessibility by allowing users to manage complex personal information, assist with educational research, and support long-horizon collaborative tasks without the prohibitive costs of current LLM-based memory architectures.

\paragraph{Potential harms and risks.} \name{} keeps every turn verbatim and, by design, never merges or discards what it stores. This preserves evidence for recall, but it also means that sensitive details persist, and fine-grained search over a person's textual and visual history could be repurposed for tracking or profiling. Append-only storage is a design choice for recall, not a privacy guarantee: a deployment needs user-initiated deletion (removing an episode from both indexes and re-summarizing its segment), retention limits, and access control. Because nothing is verified at write time, false or malicious content also persists: text injected into a conversation is stored like any other turn and can later reach the reader's prompt as retrieved evidence, a memory-injection risk shared by retrieval-based memory systems. Finally, retrieval can miss relevant memories, and the reader may then answer from partial evidence; on LongMemEval-S, \name{} declines 80\% of the unanswerable questions, so the remainder receive an unsupported answer.

\section{Limitations}
While \name{} provides promising results across the three datasets tested, and substantially reduces memory construction cost and time, we acknowledge there are some important limitations. First, due to the fact that \name{} does not rely on LLM consolidation or organization during storage, retrieving and combining memories across sessions becomes a harder task at retrieval time, leading to slightly decreased performance in questions that require these abilities. Additionally, while applying a similar approach as the text-image fusion to other modalities is feasible, the current system only handles text and image modalities.

\section{Prompt templates}
\label{sec:prompt}
In this section, we present the prompts \name{} sends to an LLM. Storage uses a single prompt, called once per conversational segment to write its summary, and retrieval uses a single prompt, called once per question to extract the answer; candidate generation, expansion, and reranking require no prompt. For completeness, we also include the prompt of the LLM judge used in evaluation. Placeholders are shown in braces.

\begin{tcolorbox}[
colback=gray!5,
colframe=blue!70,
title=\textbf{Segment Summary Prompt (Storage, One Call per Segment)},
fonttitle=\bfseries,
arc=2mm
]
\small
\textbf{Prompt:} \\
Below are consecutive excerpts from one conversation.

\{excerpts\}

Write two sentences stating what this stretch of the conversation is about and what is currently true of it. Name the people, places and things explicitly --- this will be read on its own, without the excerpts.

SUMMARY:

\medskip
\textbf{Placeholders:}
\begin{itemize}
    \item \{excerpts\}: up to $14$ evenly spaced member episodes of the segment in chronological order, one per line, each prefixed by ``- '' and truncated to $220$ characters. Decoding uses temperature $0$ and at most $200$ output tokens.
\end{itemize}
\end{tcolorbox}

\begin{tcolorbox}[
colback=gray!5,
colframe=blue!70,
title=\textbf{Answer-Extraction Prompt (Retrieval, One Call per Question)},
fonttitle=\bfseries,
arc=2mm
]
\small
\textbf{Prompt:} \\
Answer the question using the conversation excerpts below. \\
\{image\_note\}

EXCERPTS: \\
\{excerpts\}

QUESTION: \{question\}

Instructions:
\begin{itemize}
    \item Base the answer on the excerpts, and reason from them when the answer is not stated outright. If the question asks what someone would likely do, prefer, choose or be, commit to the best-supported answer instead of declining.
    \item Reply with the shortest phrase that fully answers the question. No preamble, no explanation, no restatement of the question.
    \item If the question asks for several things, include all of them.
    \item Match the wording and date style of the excerpts. Text in parentheses is a resolved value you should use as written.
    \item Reply exactly NOT\_FOUND only if the excerpts contain nothing relevant to the question. Needing to infer is not a reason to answer NOT\_FOUND.
\end{itemize}

ANSWER:

\medskip
\textbf{Placeholders:}
\begin{itemize}
    \item \{excerpts\}: the retrieved episodes and their context-window neighbors in chronological order, each rendered as ``EXCERPT $i$ (recorded $\langle$session date$\rangle$):'' followed by its text. Relative time expressions are annotated with the absolute date they denote, e.g., ``yesterday (7 May 2023)'', and an image-bearing turn carries ``[Image description: $\langle$caption$\rangle$]''.
    \item \{image\_note\}: empty in textual mode. In multimodal mode, ``The first image belongs to the question; the rest come from the conversation. Use both text and images.'' when the question carries an image, and otherwise ``Some excerpts include images. Use both the text and the images.''; either is followed by ``The images are attached in this order: 1.~$\langle$image id$\rangle$ (excerpt $i$); \ldots''.
\end{itemize}
On MemGallery, as in the benchmark's reference implementation, the benchmark's own system prompt is prepended to this prompt, and for three of its nine question categories (AR, CD, and VS) its answer-format instruction is appended to the question \citep{bei2026mem}.
\end{tcolorbox}

\begin{tcolorbox}[
colback=gray!5,
colframe=blue!70,
title=\textbf{LLM Judge Prompt (Evaluation)},
fonttitle=\bfseries,
arc=2mm
]
\small
\textbf{Prompt:} \\
You are grading an answer to a question about a long conversation.

QUESTION: \{question\} \\
CORRECT ANSWER: \{gold\} \\
STUDENT ANSWER: \{prediction\}

Mark the student answer CORRECT if it conveys the same information as the correct answer. Ignore differences in wording, phrasing, extra detail, date format, or verbosity. A correct answer that adds true context is still CORRECT.

Mark it WRONG if it states something different, contradicts the correct answer, omits a required part of it, or declines to answer.

Output ONLY JSON: \texttt{\{"verdict": "CORRECT" or "WRONG", "reason": "<few words>"\}}
\end{tcolorbox}

\section{Experiment compute resources}
All open-weight models ran on one server with four NVIDIA RTX PRO 6000 Blackwell GPUs (96\,GB each; NVIDIA driver 595.58.03, CUDA 13.2). They were served by Ollama, with the default build of each model tag (\texttt{qwen3:8b}, \texttt{gemma3:27b}, \texttt{gemma4:31b}, \texttt{qwen2.5vl:7b}, and \texttt{gpt-oss:20b} for judging). The sentence encoder and cross-encoder, both MiniLM models of about 22M parameters, ran on the same GPUs. \texttt{GPT-4.1-mini}, \texttt{GPT-4o}, and \texttt{GPT-4o-mini} were accessed through the OpenAI API.

\paragraph{Per-run cost.} Table~\ref{tab:stage_cost} gives the measured cost of one complete run on LoCoMo and MemGallery: with \texttt{Gemma3-27B}, building the memory of all ten LoCoMo conversations takes 946\,s on one GPU, and each question takes 1.95\,s with eight model instances sharing four GPUs. On this server, a full LoCoMo configuration with \texttt{Gemma3-27B} (1,540 questions) took about 8.5 minutes of wall-clock time, and a MemGallery configuration with \texttt{Qwen2.5-VL-7B} about 5 minutes as text only and 20 minutes with images. On LongMemEval-S with \texttt{Qwen3-8B}, building each history's memory took 43\,s on average, and answering took 1.1\,s per question. A LoCoMo run on \texttt{GPT-4.1-mini} or \texttt{GPT-4o} sends 2.94M input tokens and receives 10--14k output tokens.

\paragraph{Total compute.} The experiments reported in this paper, including the ablations, sensitivity studies, readers, cost measurements, and judging, took about 10.5 hours of wall-clock time on the server, or roughly 42 GPU-hours.

\end{document}